%% file: main.tex
\documentclass[10pt,twocolumn,letterpaper]{article}

\usepackage[pagenumbers]{cvpr} 


\usepackage{booktabs}
\usepackage{pifont}
\usepackage{xcolor}
\usepackage{colortbl}
\usepackage{placeins}
\usepackage{longtable}
\usepackage[table]{xcolor}
\usepackage{makecell}
\usepackage{float}
\usepackage{marvosym}  

\definecolor{cvprblue}{rgb}{0.21,0.49,0.74}
\usepackage[pagebackref,breaklinks,colorlinks,allcolors=cvprblue]{hyperref}

\definecolor{tablered}{RGB}{255,200,200}
\definecolor{tableblue}{RGB}{200,220,255}
\definecolor{tablepurple}{RGB}{222,210,222}
\definecolor{tableyellow}{RGB}{255, 255, 200}

\newcommand{\dataset}{\textbf{PercepTax}}

\def\paperID{130} 
\def\confName{CVPR}
\def\confYear{2026}
\title{Perceptual Taxonomy: Evaluating and Guiding Hierarchical Scene Reasoning in Vision-Language Models}

\author{
Jonathan Lee\textsuperscript{1*} \quad
Xingrui Wang\textsuperscript{1*} \quad
Jiawei Peng\textsuperscript{1} \quad
Luoxin Ye\textsuperscript{1} \quad
Zehan Zheng\textsuperscript{1} \quad \\
Tiezheng Zhang\textsuperscript{1} \quad
Tao Wang\textsuperscript{1} \quad
Wufei Ma\textsuperscript{1} \quad
Siyi Chen\textsuperscript{1} \quad \\
Yu-Cheng Chou\textsuperscript{1} \quad
Prakhar Kaushik\textsuperscript{1$\dagger$\Letter}  \quad
Alan Yuille\textsuperscript{1$\dagger$} \\
\textsuperscript{1}Johns Hopkins University 
}

\usepackage{multirow}
\usepackage{multicol}
\usepackage{pifont}

\definecolor{deepgreen}{RGB}{0,100,0} 
\newcommand{\cmark}{\textcolor{deepgreen}{\checkmark}} 
\definecolor{deepred}{RGB}{189,20,0} \newcommand{\xmark}{\textcolor{deepred}{\ding{55}}}

\begin{document}

\twocolumn[{%
\renewcommand\twocolumn[1][]{#1}%
\maketitle
\vspace{-2.5em}
\centering
\url{https://perceptual-taxonomy.github.io/}

\begin{center}

\includegraphics[width=0.95\linewidth]{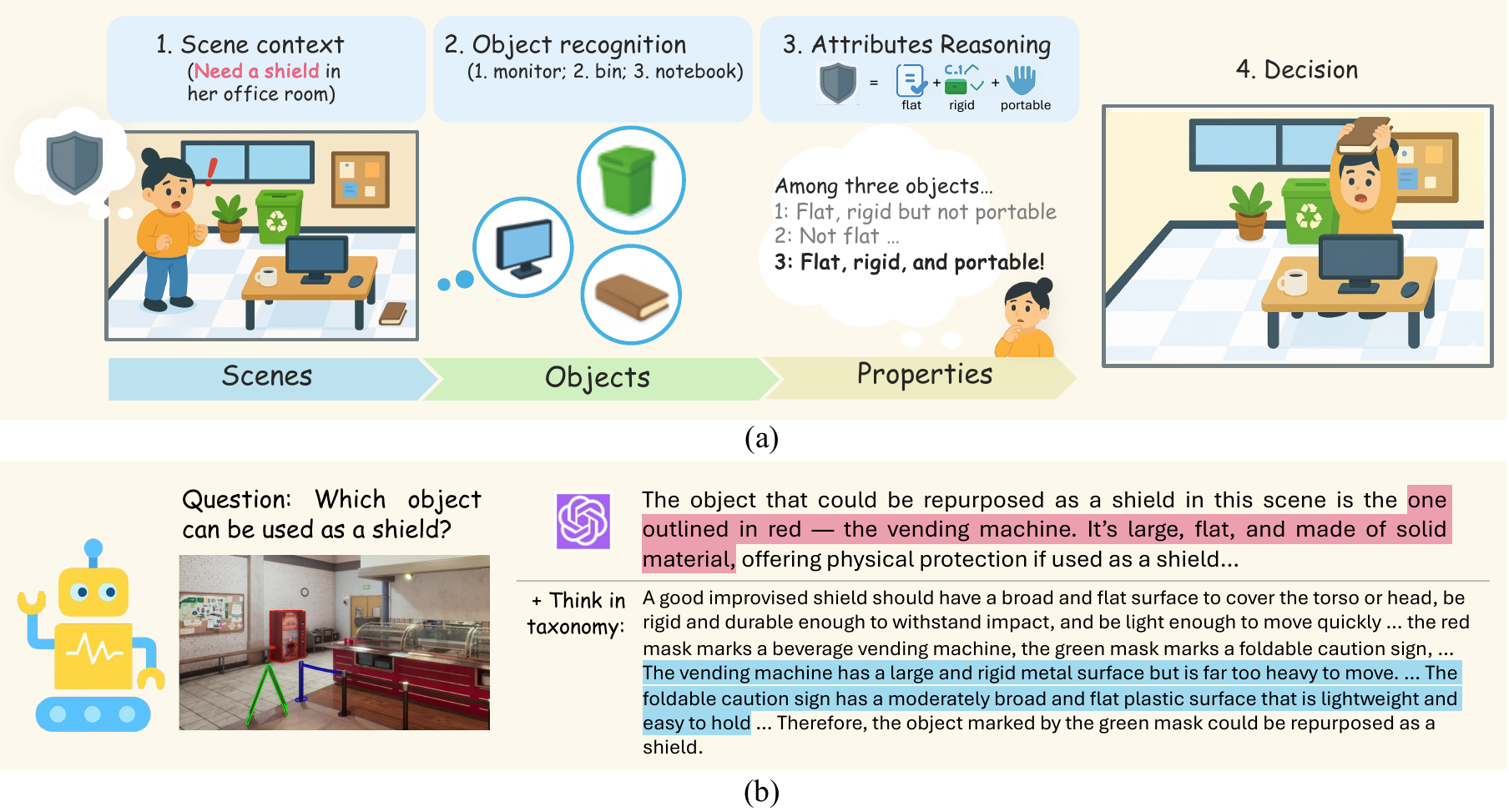}
\captionof{figure}{
\textbf{Illustration of our proposed \emph{Perceptual Taxonomy} for structured scene understanding} 
(a) A motivating example of human reasoning: when needing protection, a person hierarchically perceives the scene (scene-level context), identifies available objects (object-level), infers their physically grounded properties (e.g., flat, rigid, portable), and selects the most suitable one for the goal (book as shield). 
(b) An example from \dataset{} where a model is asked to select an object that can be repurposed as a shield. The model must reason over object properties to make a functional choice, emulating human-like perception and decision-making.\vspace{-1em}}
\label{fig:Taxonomy}
\end{center}
\vspace{1em}
}]

\begingroup
\renewcommand\thefootnote{}%
\footnotetext{%
\textsuperscript{*} Equal contribution. \quad
\textsuperscript{$\dagger$} Joint senior authors}
\footnotetext{\textsuperscript{\Letter} Corresponding author: Prakhar Kaushik (pkaushi1@jhu.edu)}

\addtocounter{footnote}{-1}%
\endgroup


\input{sec/0_abstract}

\input{sec/1_intro}
\input{sec/2_related_work}

\input{sec/3_benchmark}

\input{sec/4_data_curation}

\input{sec/5_experiments}

\input{sec/6_conclusion}

\clearpage

\newpage

{
    \small
    \bibliographystyle{ieeenat_fullname}
    \bibliography{main}
}

\appendix
\input{sec/7_suppl}

\end{document}

%% file: sec/0_abstract.tex
\begin{abstract}


We propose Perceptual Taxonomy, a structured process of scene understanding that first recognizes objects and their spatial configurations, then infers task-relevant properties, such as material, affordance, and function to support goal-directed reasoning. While this form of reasoning is fundamental to human cognition, current vision language benchmarks lack comprehensive evaluation of this ability, focusing instead on surface-level recognition or image text alignment.
To address this gap, we introduce \dataset, a benchmark for physically grounded visual reasoning. We annotate \textbf{3,173} objects with four property families: material, physical attributes, affordance, and function  covering \textbf{84} fine-grained attributes. Based on these annotations, we construct a multiple-choice question (MCQ) benchmark with \textbf{5,802} images spanning both synthetic and real domains. It includes \textbf{28,033} template based questions spanning four types (object description, spatial reasoning, property matching, and taxonomy reasoning), as well as \textbf{50} expert-crafted questions, designed to evaluate models across the full spectrum of perceptual taxonomy reasoning.
Experimental results reveal that leading VLMs perform well on recognition tasks but degrade significantly by \textbf{10 - 20\%} on property-driven questions, especially those requiring multi-step reasoning over structured attributes. These findings highlight a persistent gap in structured visual understanding and the limitations of current VLMs reliance on pattern matching. Further, we show that providing in-context reasoning exemplars from simulation data improves performance on real-world and expert-curated questions, demonstrating the effectiveness of perceptual taxonomy guided prompting.
\vspace{-1em}
\end{abstract}

%% file: sec/1_intro.tex
\section{Introduction}

\label{sec:intro}


Human perception and reasoning follow a \emph{Perceptual Taxonomy} -- a hierarchical process of extracting structured information from visual input~\cite{adams2015bloom,shaw2017perceiving}. We first recognize objects and their spatial configurations, then infer task-relevant properties such as material, affordance, and function.
These perceptual properties underlie how humans interact with the physical world~\cite{yuille2006vision,gibson1979ecological,greene2016visual}. As illustrated in \cref{fig:Taxonomy}(a), when a person seeks protection but no shield is available, they naturally survey the environment (e.g., monitor, bin, notebook), infer which properties are required (e.g., flat, rigid, portable), and select the object that best satisfies the goal -- such as repurposing a book as a shield.

We formalize this process as \textbf{\emph{Perceptual Taxonomy}}, a hierarchical framework that (1) \emph{perceives}, by decomposing visual scenes into objects and their physically grounded properties; then (2) \emph{reasons}, by using these properties to support goal-directed inference, such as identifying affordances, functions, or materials relevant to a task.

Although vision–language models (VLMs) have become the leading systems for multimodal reasoning~\citep{comanici2025gemini,claude,gpt5,qwen3technicalreport,wang2025internvl3_5,liu2024llavanext}, most existing benchmarks still emphasize surface-level recognition or text–image alignment, offering limited insight into whether models can reason over \emph{perceptual taxonomy}. 
As summarized in~\cref{tab:benchmark_comparison}, early benchmarks such as CLEVR, GQA, and VQA v2.0~\cite{goyal2017making,johnson2017clevr,hudson2019gqa} laid the groundwork for compositional and relational reasoning, while more recent datasets~\cite{azuma2022ScanQA,li2023seed,liu2024mmbench,tong2024cambrian,wang2025spatial457,wu2024star} extend to common sense and causal reasoning. 
Despite this progress, these benchmarks primarily assess high-level semantic understanding and neglect physically meaningful attributes such as material, affordance, and function. 
Consequently, they fail to evaluate whether models can bridge perceptual grounding with hierarchical reasoning—an ability essential for understanding and interacting with real-world scenes. 
Most importantly, current VLMs still lack \emph{perceptual taxonomy}: the capacity to reason over object properties and use them to guide inference in novel scenarios.

\begin{table*}[t]
\centering

\caption{
Comparison of benchmarks across scene- and object-level reasoning dimensions. 
Existing visual question answering and reasoning benchmarks primarily focus on 2D/3D scene understanding or basic object descriptions, while only a few address property-level reasoning such as material or affordance. 
None systematically cover hierarchical \textit{perceptual taxonomy}. 
Our proposed benchmark (bottom row) uniquely spans both simulated and real images, providing comprehensive coverage across all perceptual dimensions.
}

\resizebox{\textwidth}{!}{
\begin{tabular}{lccccccccccc}
\toprule
\multirow{2}{*}{\textbf{Benchmarks}} & \multirow{2}{*}{\textbf{\#QA}} & \multirow{2}{*}{\textbf{Synth/Real}}
& \multicolumn{2}{c}{\textbf{Scene Level}}
& \textbf{Object} 
& \multicolumn{4}{c}{\textbf{Properties}}
& \textbf{Perceptual} \\
\cmidrule(lr){4-5} \cmidrule(lr){6-6} \cmidrule(lr){7-10}
& & 
& 2D & 3D
& Obj. Desc.
& Material & Affordance & Function & Phys. Props.
& \textbf{Taxonomy} \\
\midrule

CLEVR~\citep{johnson2017clevr}        
& 853K  & Synth
& \cmark & \xmark
& \cmark 
& \xmark & \xmark & \xmark & \xmark
& \xmark \\

GQA~\citep{hudson2019gqa}          
& 22M   & Real
& \cmark & \xmark
& \cmark
& \cmark & \xmark & \xmark & \xmark
& \xmark \\

VQA v2.0~\citep{goyal2017making}     
& 1.1M  & Real
& \cmark & \xmark
& \cmark
& \xmark & \xmark & \xmark & \xmark
& \xmark \\


ScanQA~\citep{azuma2022ScanQA}
& 41k    & Synth
& \cmark & \cmark
& \cmark
& \xmark & \xmark & \xmark & \xmark
& \xmark \\

SEED-Bench-2~\citep{li2023seed}  
& 24K    & Real
& \cmark & \xmark
& \cmark
& \cmark & \xmark & \xmark & \xmark
& \xmark \\

MMBench~\citep{liu2024mmbench}          
& 3.2K    & Real
& \cmark & \xmark
& \cmark
& \xmark & \xmark & \xmark & \xmark
& \xmark \\

CV-Bench~\citep{tong2024cambrian}          
& 2.6K    & Real
& \cmark & \cmark
& \cmark
& \xmark & \xmark & \xmark & \xmark
& \xmark \\

Spatial457~\citep{wang2025spatial457}  
& 24K    & Synth
& \cmark & \cmark
& \cmark
& \xmark & \xmark & \xmark & \xmark
& \xmark \\

\midrule

RIO~\citep{qu2023rio}  
& 130K    & Real
& \cmark & \xmark
& \cmark
& \xmark & \cmark & \xmark & \xmark
& \xmark \\

Robo2VLM~\citep{chen2025robo2vlm}  
& 685K    & Real
& \cmark & \cmark
& \cmark
& \xmark & \cmark & \xmark & \xmark
& \xmark \\

RoboAfford~\citep{tang2025roboafford}  
& 1.9M    & Synth+Real
& \cmark & \cmark
& \cmark
& \xmark & \cmark & \xmark & \xmark
& \xmark \\

A4Bench~\citep{wang2025affordance}  
& 1,282    & Real
& \xmark & \xmark
& \cmark
& \xmark & \cmark & \xmark & \xmark
& \xmark \\

PTR~\citep{hong2021ptr}           
& 700K & Synth
& \cmark & \cmark
& \cmark
& \xmark & \xmark & \xmark & \cmark
& \xmark \\

PhysGame~\citep{cao2024physgame}  
& 880    & Real
& \cmark & \cmark
& \cmark
& \cmark & \xmark & \xmark & \cmark
& \xmark \\

PhysBench~\citep{chow2025physbench}
& 10k   & Real
& \cmark & \xmark
& \cmark
& \xmark & \xmark & \xmark & \cmark
& \xmark \\

Hypo3D~\citep{mao2025hypo3d}  
& 15k    & Synth
& \cmark & \cmark
& \cmark
& \cmark & \xmark & \cmark & \xmark
& \xmark \\


\midrule
\dataset{}   
& 28K    & Synth + Real
& \cmark & \cmark
& \cmark
& \cmark & \cmark & \cmark & \cmark
& \cmark \\


\bottomrule
\end{tabular}
}
\label{tab:benchmark_comparison}
\end{table*}

Benchmarking \emph{perceptual taxonomy} reasoning is non-trivial due to two key limitations:
(i) the lack of a carefully designed, large-scale annotation scheme that defines a shared set of object properties—such as material, affordance, functionality, and physical attributes, and cluster common object categories to match with the properties;
(ii) the absence of diverse and comprehensive question types and questions templates that require reasoning over these properties in varied contexts, an essential reasoning ability that current VLM benchmarks fail to evaluate. 


We propose \textbf{\dataset{}} as the first benchmark to systematically evaluate \emph{perceptual taxonomy} reasoning in vision language models. This framework reflects how humans interpret scenes by first identifying objects and spatial layouts, then inferring task-relevant properties such as material, affordance, and function. Existing benchmarks lack comprehensive evaluation of this reasoning ability, focusing instead on visual matching or recognition. To address this gap, we construct a standardized property dictionary spanning four core physical attributes and annotate \textbf{3,173} object classes with \textbf{84} property types. Each scene is represented as a hierarchy from scene to object to property, and grounded with names, 3D positions, and structured annotations. The dataset includes \textbf{5,802} images from both simulated and real domains, each paired with property labels. Building on this foundation, we design four template based question, includes object description, spatial reasoning, property matching, and taxonomy reasoning, yielding a total of \textbf{28,033} multiple-choice questions. We also includes \textbf{50} expert hancrafted real-image queries. Our evaluation shows that leading models often struggle with property-level inference, especially on taxonomy-driven questions. This highlights a reliance on pattern matching rather than structured understanding, motivating future development of models with stronger perceptual reasoning capabilities.

In our experiments, we evaluate state-of-the-art closed-source (e.g., Gemini 2.5~\cite{comanici2025gemini}, GPT-5~\cite{gpt5}, Claude Sonnet 4.5~\cite{claude}) and open-source (e.g., Qwen3-VL~\cite{qwen3technicalreport}, InternVL~\cite{wang2025internvl3_5}, LLaVA-Next~\cite{liu2024llavanext}) VLMs on \dataset{}.
While these models perform strongly on object recognition—for instance, Gemini~2.5 and GPT-5 achieve {92.16\%} and {88.63\%} on real-image object description—their accuracy drops sharply at the property level and, most notably, in taxonomy reasoning. 
Even GPT-5 reaches only {58.85\%} on real-image taxonomy reasoning and {57.59\%} on simulation, representing a drop of more than 29 points from its object-level performance, indicating that current VLMs struggle to integrate material, affordance, and functionality cues into coherent physical understanding.

To further validate the necessity of perceptual taxonomy in reasoning, we design a reasoning paradigm aligned with the hierarchical structure of \dataset{}. 
In this paradigm, the model is guided to reason over the question by traversing the scene hierarchy, from the scene layout, to relevant objects, and finally to their properties. 
During experiments, we adopt a few-shot adaptation strategy using samples from the simulation data to provide structured exemplars of object property relationships. 
This taxonomic reasoning process substantially improves the model's performance on real-world images, demonstrating that learning hierarchical physical structures from simulation can generalize to realistic settings.

Our contributions can be summarized as follows:  

\begin{enumerate}
    \item We formalize \emph{perceptual taxonomy} and construct a comprehensive property dictionary, annotating \textbf{84} physics-relevant attributes across \textbf{3,173} object classes, including material, affordance, function, and physical properties.

    \item We propose \dataset{}, a visual question answering benchmark containing \textbf{5,802} images (simulated and real) and \textbf{28,083} questions across four task families: object description, spatial reasoning, property matching, taxonomy reasoning with 50 expert handcrafted questions.

    \item We benchmark state-of-the-art VLMs and reveal substantial weaknesses in property-level and taxonomy reasoning.  
    We further propose a taxonomic reasoning paradigm that improves performance via few-shot in-context learning, even enabling sim-to-real transfer, highlighting the importance of perceptual taxonomy for structured physical reasoning.
\end{enumerate}

%% file: sec/2_related_work.tex
\vspace{-0.5em}
\section{Related Work}
\label{sec:related}
\subsection{VLM benchmark for scene understanding}
Recent advances in large-scale vision and language models (VLMs) have shown strong generalization across diverse multimodal tasks~\cite{yang2025thinking,kil2024mllm,yue2024mmmu,fu2024blink}. Early benchmarks such as VQA v2.0~\cite{goyal2017making}, CLEVR~\cite{johnson2017clevr}, and GQA~\cite{hudson2019gqa} established foundations for compositional reasoning, focusing on object attributes, spatial relations, and logical consistency. Later datasets including NLVR2~\cite{suhr2019nlvr2}, OK-VQA~\cite{marino2019okvqa}, VCR~\cite{zellers2019vcr}, and A-OKVQA~\cite{schwenk2022aokvqa} introduced language grounded and commonsense reasoning challenges.
Recent works such as Winoground~\cite{thrush2022winoground}, PTR~\cite{hong2021ptr}, and SuperCLEVR~\cite{li2023superclevr} target fine grained compositionality and distribution shifts. Datasets like ScanQA~\cite{azuma2022ScanQA} and Spatial457~\cite{wang2025spatial457} expand into 3D spatial reasoning. Meanwhile, general purpose evaluation suites such as MMBench~\cite{liu2024mmbench}, SEED-Bench-2~\cite{li2023seed}, and CV-Bench~\cite{tong2024cambrian} assess a broad range of multimodal skills.
While these benchmarks advance structured multimodal reasoning, they primarily focus on semantics or geometry. A systematic framework for evaluating reasoning over attributes such as material, affordance, and function remains underexplored, yet it is essential for grounding VLMs in real world understanding.

\begin{figure*}[t]
\centering
\includegraphics[width=0.95\linewidth]{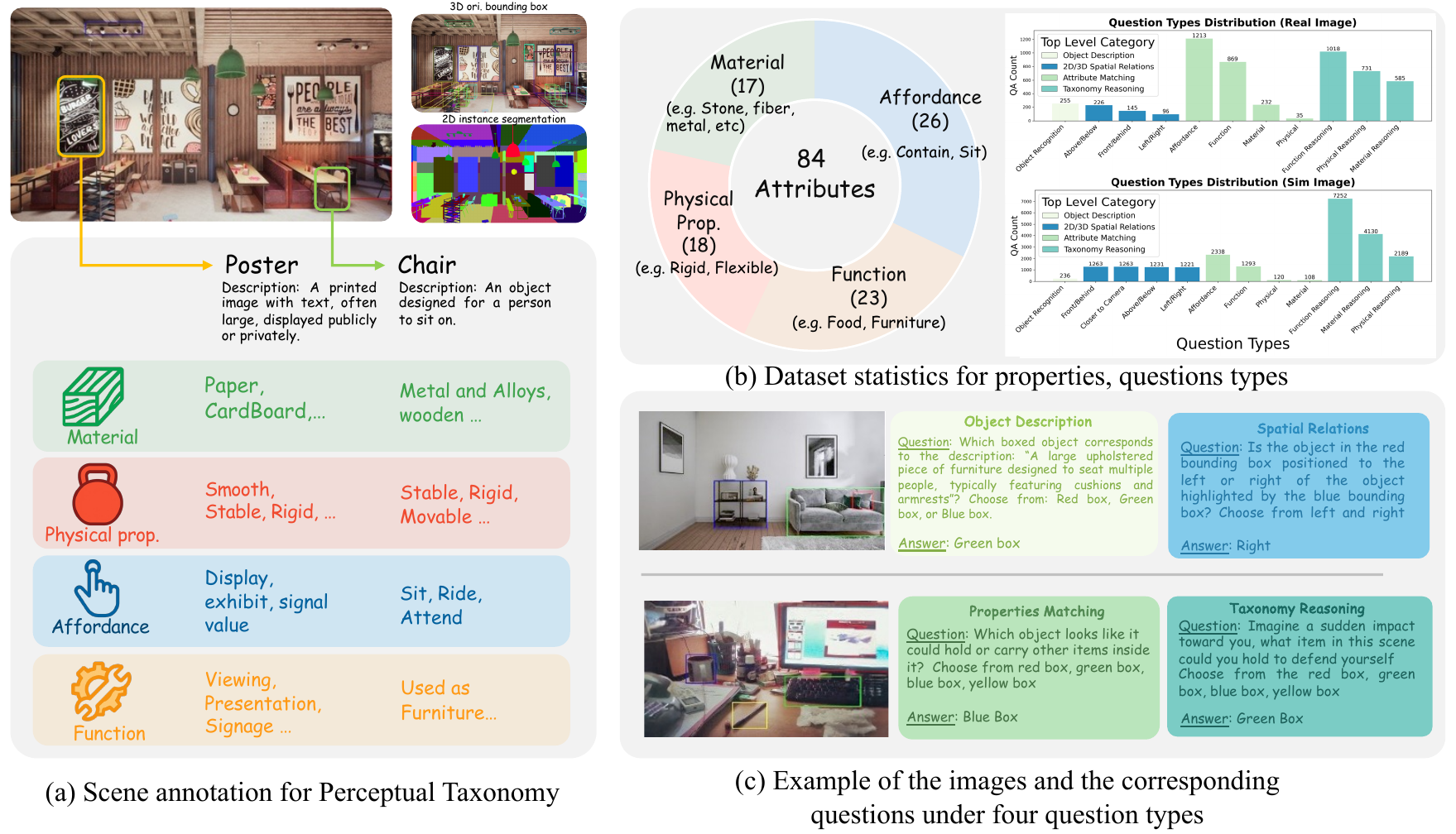}
\caption{
Overview of the \dataset{} benchmark.
(a) Each scene is annotated with object detections and their mapped attributes across four domains: material, physical properties, affordance, and function.
(b) Summary statistics for 84 attribute categories and question distributions across synthetic and real images.
(c) Example questions covering four reasoning types: object description, spatial relation, property matching, and taxonomy reasoning, together forming a unified framework for perceptual and hierarchical scene understanding.\vspace{-1em}}
\label{fig:benchmark}

\end{figure*}

\subsection{Object Properties Reasoning}

Beyond general scene understanding, several works focus on reasoning over object properties such as material, affordance, functionality, and physical attributes. ShapeStacks~\cite{groth2018shapestacks} and Falling Tower~\cite{Balazadeh2025fallingtower} examine intuitive physics through stability prediction in stacking scenarios. PhysGame~\cite{cao2024physgame} challenges models to detect physically implausible scenes from gameplay videos, emphasizing material and broader physical reasoning. PhysBench~\cite{chow2025physbench} introduces detailed property annotations in real world images, while PhysXNet~\cite{cao2025physx} provides large scale 3D assets to support VQA on physical properties.

ManipVQA~\cite{huang2024manipvqa} frames affordance and tool-use understanding as a VQA task, encouraging models to reason about object manipulation. A4Bench~\cite{wang2025affordance} further explores constitutive and transformative affordances, connecting static and dynamic object interaction.

While these benchmarks offer valuable insights into specific aspects of physical reasoning, most evaluate isolated attributes in limited contexts. They often lack integration of scene level structure, object recognition, and property inference. Our work addresses this gap by introducing a unified benchmark that evaluates perceptual taxonomy reasoning through structured object property annotation and scene contextual tasks.


%% file: sec/3_benchmark.tex
\vspace{-1em}
\section{\dataset{}: A Benchmark for Perceptual Taxonomy Reasoning}
\vspace{-0.5em}
\label{sec:benchmark}

\subsection{Perceptual Taxonomy and Property Definition}
\label{sec:scene_structure}
To evaluate physically grounded scene understanding, we define a hierarchical representation that mirrors how humans organize visual information. As illustrated in \cref{fig:benchmark} (a), each scene in \dataset{} follows a three-level hierarchy:
\emph{scene} $\rightarrow$ \emph{objects} $\rightarrow$ \emph{properties}.
(1) For each object, we annotate a 2D instance segmentation mask for accurate image-plane localization (together with a 3D oriented bounding box) capturing its position, depth, and yaw in the camera coordinate frame to support explicit spatial and geometric reasoning within the scene.
(2) At the \emph{property level}, each object is annotated with a comprehensive set of attributes capturing physically meaningful semantics:

\begin{itemize}[leftmargin=1.2em]
    \item \textbf{Material}: the object's visual and physical substance (e.g., paper, metal, wood), which determines its rigidity, weight, texture, and durability.
    
    \item \textbf{Physical properties}: intrinsic physical characteristics that govern how the object behaves under external forces, such as rigidity, fragility, elasticity, mass, or stability.
    
    \item \textbf{Affordance}: the set of actions or interactions the object enables based on its shape and components (e.g., graspable, supportable, sit-on, containable), independent of its intended purpose.
    
    \item \textbf{Functionality}: the object's intended or contextual role when invented (e.g., seating furniture, storage container, display signage), reflecting how humans typically use it in practice.
\end{itemize}
As shown in \cref{fig:benchmark}(a), a restaurant scene may include objects such as a \emph{poster} and a \emph{chair}. Each object is grounded by its canonical description and then linked to its four families of properties—for example, a poster (\emph{paper}, \emph{smooth}, \emph{display}, \emph{signage}) versus a chair (\emph{wood and metal}, \emph{stable}, \emph{sit}, \emph{furniture}). 
We define a taxonomy covering \textbf{16} material types, \textbf{43} affordances, \textbf{23} functionalities, and \textbf{17} physical properties, as illustrated in \cref{fig:benchmark}(b).
A total of \textbf{3,173} unique objects are annotated, with many-to-many mappings to these properties.
Full list of category-to-object mappings are provided in the Appendix.
These structural annotations establish the foundation for \emph{perceptual taxonomy} reasoning tasks, which require understanding spatial position or reasoning over structured physical meaningful properties.


\subsection{Template-based Question and Generation}
\label{sec:task_types}
We begin by leveraging the comprehensive annotations described in Section~\ref{sec:scene_structure}, including per-object material, affordance, function, physical properties, and 2D / 3D spatial cues. Building on the engine from CLEVR~\citep{johnson2017clevr}, we design 48 parameterized templates spanning four core perceptual taxonomy tasks: (1) spatial understanding, (2) object description, (3) attribute matching, and (4) taxonomy reasoning.
We describe each task category below and we list all the templates in the Appendix.

\textbf{(1) Object Descriptions} questions target the most basic layer of perceptual taxonomy: matching a natural-language description from an object's material, physical properties, affordances, or functional cues to the correct color coded bounding box. These questions focus on direct property object alignment rather than multi-step reasoning. For example, “Which object marked by the colored boxes is used for writing with chalk?” requires selecting the chalkboard based on its functional property, also as the example illustrated in \cref{fig:benchmark}(c).

\textbf{(2) Spatial Reasoning} forms another core component of perceptual taxonomy, capturing how objects are arranged and interact within a scene. Prior work has explored both 2D and 3D spatial reasoning, and consistently shows that VLMs remain weak—particularly in inferring accurate 3D spatial relations~\cite{johnson2017clevr,goyal2017making,hudson2019gqa,cheng2024spatialrgpt,wang2025spatial457,armeni20193d,ma20253dsrbench,ma2025spatialreasoner}. Motivated by these findings, we include spatial reasoning as a key component of perceptual scene understanding.

We follow similar question designs and generate prompts that ask models to identify relations such as left/right, above/below, and front/behind between color-coded objects. All relations are computed directly from ground-truth 2D image coordinates or 3D object locations in the camera frame (see \cref{sec:data_curation} for details). An example question is: “Is the object marked by the Green box to the left or right of the object marked by the Blue box?” Additional instances are shown in \cref{fig:benchmark}(c).

\textbf{(3) Attribute Matching} questions evaluate whether a model can associate specific physical or functional attributes with the correct object in the scene. All queried attributes—such as material, rigidity, elasticity, thermal properties, or functional affordances—are drawn directly from our taxonomy-based annotations (as in Section~\ref{sec:scene_structure}).

Unlike object-description questions, which focus on coarse, appearance-level cues, attribute matching requires models to go beyond surface visual features and recognize properties that reflect real-world physical semantics. Correctly answering these questions often requires understanding that certain attributes (e.g., “rigid,” “conductive,” “container-like,” “made of slate”) cannot be inferred purely from color or shape, but must be grounded in how objects behave or are used in the physical world.

For example, in the question “Which object marked by the colored boxes is made of slate?”, the correct answer is the chalkboard. Although visually similar distractors exist, they do not share the underlying material property. Such questions therefore test the model’s ability to distinguish fine-grained, attribute-level differences and ground them in both visual evidence and physical-world knowledge, rather than superficial appearance alone.

\FloatBarrier
\begin{figure*}[!h]
\centering
\includegraphics[width=0.8\linewidth]{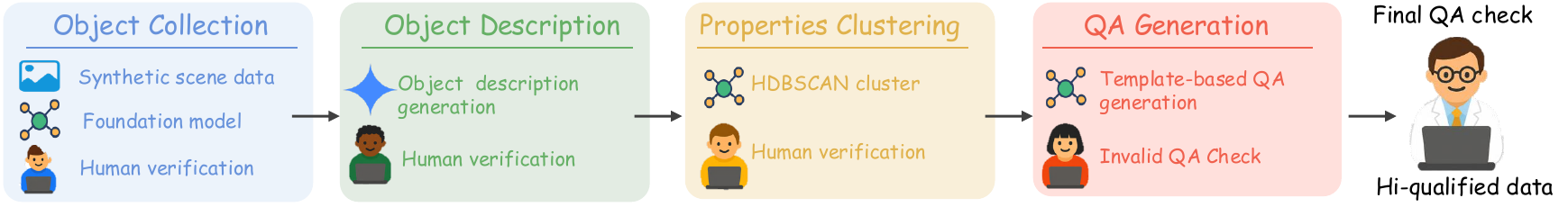}
\caption{Object Anotation Pipeline: 
Objects are first collected from synthetic and real scenes, described by foundation models, and clustered by shared material, function, and affordance attributes with human verification at each stage. 
Template-based QA generation then converts verified clusters into question–answer pairs, with 68.3\% passing the final quality check to ensure accurate and semantically grounded annotations.
}
\label{fig:Annotation}
\end{figure*}

\textbf{(4) Taxonomy Reasoning} questions represent the highest level of perceptual taxonomy in question template and are the primary focus of our benchmark. Unlike previous categories that involve direct property--object alignment, these questions require multi-step inference over multiple structured attributes, often combining affordances, material properties, physical constraints, and functional compatibility. To answer correctly, a model must perform a filter--eliminate reasoning process: identifying the attributes implied by the goal, inferring which objects possess the necessary physical or functional properties, and ruling out candidates that fail any of the constraints. This goes beyond appearance recognition and instead demands a grounded understanding of how objects behave and can be repurposed in the physical world. For example, the question \emph{``Which object could function as a temporary stepstool?''} requires recognizing that the correct choice must be rigid, stable, and provide a flat, load-bearing surface at an appropriate height—properties only one object in the scene satisfies. Such questions probe whether models can integrate multiple dimensions of perceptual taxonomy to reach coherent and physically plausible conclusions.

With these templates defined, our question engine instantiates large numbers of questions across scenes. We apply lightweight LLM-based~\cite{comanici2025gemini} rephrasing to increase linguistic variety, while a series of automatic validity checks ensures that each question is grounded, unambiguous, and aligned with the intended attribute or relation. These checks remove objects or scenes that would introduce semantic noise (e.g., unsuitable attribute clusters or multiple valid answers), ensuring that every retained question has a single, clearly identifiable solution. Together with brief human verification, this process yields a diverse yet reliable set of questions that directly reflect the structured annotations and taxonomy in our benchmark. All operational details are provided in the Appendix.


\subsection{Handcrafted Human-Level Reasoning}
Beyond the template-based questions, we include a small set of human-crafted reasoning questions to evaluate abilities that require integrated, hierarchical understanding of a scene—such as combining material, functionality, affordances, spatial relations, and counterfactual effects. Unlike prior human-level benchmarks such as ZeroBench~\cite{roberts2025zerobench}, which often rely on open-ended judgments or external knowledge, our questions are fully visually grounded and objectively answerable, targeting reasoning skills that template expansion alone cannot capture. To construct this subset, human experts curated questions from a 50-image sample of our Real Image Set, designing one question per image to ensure meaningful, physically plausible reasoning grounded in the scene. Further details and examples are provided in the Appendix.



%% file: sec/4_data_curation.tex

\subsection{Data Curation}
\label{sec:data_curation}
\noindent\textbf{Simulation image rendering.}
We generate our synthetic image set using 33 high-quality Unreal Engine~5~\cite{unrealengine5} assets, which provide accurate geometry, physically based materials, and photorealistic lighting. This ensures that all rendered images come with precise and complete scene annotations, including 2D instance masks, 2D/3D bounding boxes, object poses, and material and physical attributes. 
To obtain diverse and semantically rich images, we randomly sample camera viewpoints from fully constructed scenes to capture realistic layouts and natural spatial relationships. We further augment scene diversity by modifying environments: all placeable objects are manually curated and randomly arranged on valid surfaces (e.g., tables, floors) to produce additional compositions and layouts. These pipelines yield a 4,544 realistic and well-annotated images. The operational details of rendering, camera sampling, and quality filtering are provided in Appendix.


\noindent\textbf{Real images collection and annotation.}
We curate real-world scenes from the OpenImages~\cite{kuznetsova2020openimages} training subset. Since these images do not contain scene-level annotations, we apply several foundation models to detect objects and extract their spatial information, including 2D bounding boxes, 3D bounding boxes, and pose estimates~\cite{yang2024ramplusplus,liu2023groundingdino,oquab2023dinov2}. This allows us to obtain structured scene representations and generate 1,258 images. 


\noindent\textbf{Object description and properties clustering.}
As shown in \cref{fig:Annotation}, starting from the curated objects collected from both simulation scenes and real images, we use Gemini to generate concise textual descriptions covering visual characteristics and taxonomy-relevant attributes such as material, affordance, physical properties, and functionality. These descriptions are then grouped using unsupervised clustering methods (e.g., HDBSCAN~\cite{mcinnes2017hdbscan}, \textit{k}-means~\cite{macqueen1967kmeans}) to form initial property clusters. Human annotators review and refine these clusters to ensure semantic correctness and resolve inconsistencies, producing a reliable taxonomy that supports our QA generation pipeline. The verified clusters are then used to instantiate template-based questions grounded in the annotated scenes. 

\noindent\textbf{Final human quality verification.}
We conduct a global human quality verification stage in which annotators inspect generated QA pairs and remove problematic cases—such as inaccurate bounding boxes, incorrect object identities, or mismatched attributes with image input. Approximately 31.7\% of automatically generated questions were discarded through this process, leaving only high-quality, visually grounded QA pairs for inclusion in the benchmark.

\FloatBarrier
\begin{table*}[!ht]
\centering
\caption{
Performance of representative \textbf{close-source} and \textbf{open-source} VLMs on the 
\textbf{Simulation Set}, \textbf{Real Image Set}, and \textbf{Expert Crafted Question} subset and higher is better. 
\textbf{S.R.} = Spatial Relation; 
\textbf{Obj. Desc.} = Object Description; 
\textbf{Tax. Match.} = Attribute Matching; 
\textbf{Tax. Reason.} = Taxonomy Reasoning.
The highest score among all models for each question type is marked in \colorbox{tablered}{red}, and the second-highest scores are \underline{underlined}. The best performance among open-source models is marked in \colorbox{tableblue}{blue}
}
\label{tab:sim_real_hard_full}
\resizebox{\textwidth}{!}{
\begin{tabular}{l|ccccc|ccccc|c}
\toprule
\multirow{2}{*}{\textbf{Model}} 
& \multicolumn{5}{c|}{\textbf{Simulation Set}} 
& \multicolumn{5}{c|}{\textbf{Real Image Set}} 
& \multirow{2}{*}{\textbf{Expert}} \\ 
\cmidrule(lr){2-6} \cmidrule(lr){7-11}
& \textbf{S.R.} & \textbf{Obj. Desc.} & \textbf{Attri. Match.} & \textbf{Tax. Reason.} & \textbf{All} 
& \textbf{S.R.} & \textbf{Obj. Desc.} & \textbf{Tax. Match.} & \textbf{Tax. Reason.} & \textbf{All} 
&  \\    
\midrule
\multicolumn{12}{c}{\textit{Close-source models}} \\
\midrule
Gemini 2.5 Pro~\citep{comanici2025gemini} & \underline{66.72} & \cellcolor{tablered} 59.63 & \cellcolor{tablered} 50.44 & \cellcolor{tablered} 52.93 & \cellcolor{tablered} 55.71 & 73.32 & \cellcolor{tablered} 92.16 & \cellcolor{tablered} 81.01 & \cellcolor{tablered} 63.40 &  \cellcolor{tablered} 77.97 & 14 \\
Claude Sonnet 4.5~\citep{claude} & 56.03 & 28.39 & 32.80 & 30.93 & 36.78 & 48.80 & 65.21 & 52.32 & 40.26 & 51.65 & 10 \\
GPT-5~\citep{gpt5} & 66.09 & \underline{51.69} & 42.27 & \underline{45.78} & \underline{49.63} & \cellcolor{tablered} 74.58 & \underline{88.63} & \underline{78.42} & \underline{59.18} & \underline{70.70} &  \underline{20} \\

\midrule
\multicolumn{12}{c}{\textit{Open-source models}} \\
\midrule
Qwen3-VL-32B~\citep{qwen3technicalreport} & \cellcolor{tablered} 67.16 & \cellcolor{tableblue} 47.88 & \cellcolor{tableblue} \underline{43.54} &\cellcolor{tableblue}  43.56 & \cellcolor{tableblue} 48.79 & \cellcolor{tableblue} \underline{74.30} &  84.31 & \cellcolor{tableblue} 74.19 & \cellcolor{tableblue} 48.78 & \cellcolor{tableblue} 64.07 & \cellcolor{tablered} 24\\
InternVL3.5-30B-A3B~\citep{wang2025internvl3_5} & 38.27 & 30.01 & 29.55 & 28.98 & 31.14 & 72.16 & 83.13 & 67.12 & 37.97 & 56.15 & 16 \\
\midrule
Qwen3-VL-8B~\citep{qwen3technicalreport} & 66.21 & 44.91 & 41.57 & 38.58 & 45.29 & 73.02 & \cellcolor{tableblue} 85.10 & 69.44 & 45.26 & 60.04 & \cellcolor{tablered} 24\\
InternVL3.5-8B~\citep{wang2025internvl3_5} & 33.49 & 29.66 & 27.65 & 28.78 & 29.61 & 69.59 & 85.09 & 62.78 & 37.75 & 53.98 & 8 \\
LLaVA-OneVision-7B~\citep{liu2024llavanext} & 49.76 & 41.10  & 27.98  & 29.04  & 33.50  & 40.06  & 45.27 & 37.92  & 37.16  & 36.55  & 8 \\
LLaVA-Next-Llama3~\citep{liu2024llavanext} & 34.27 & 40.25 & 24.40 & 27.39 & 28.46 & 57.17 & 39.22 & 35.86 & 25.39 & 33.49 & 6\\
LLaVA-Next-Vicuna-7B~\citep{liu2024llavanext} & 33.96 & 49.57 & 26.13 & 27.08 & 28.64 & 55.67 & 28.24 & 28.55 & 22.45 & 28.34 & 6\\


\bottomrule
\end{tabular}}
\label{tab:main_results}
\end{table*}


\subsection{Benchmark Statistics} \label{sec:statistics}

Our benchmark consists of \textbf{5,802} images, including \textbf{4,544} synthetic images and \textbf{1,258} real images. Across these scenes, we annotated \textbf{3,173} distinct object classes, each matched many-to-many to 84 attribute labels spanning four property types: material, physical properties, affordances, and functions. In total, we generate \textbf{28,033} template-based MCQ questions covering the four core reasoning tasks introduced in Section~\ref{sec:task_types}. The distribution of question types is shown in the histogram of \cref{fig:benchmark}(b), and detailed statistics across categories and subtypes are provided in Appendix. In addition, we include \textbf{50} human-authored questions designed to test higher-level reasoning on the same real image setting.


%% file: sec/5_experiments.tex
\vspace{-0.5em}
\section{Experiments}
\label{sec:exp}

\subsection{Baselines}
We evaluate our benchmark using a diverse suite of state-of-the-art Vision Language Models (VLMs) using the open-source toolkit~\citep{duan2024vlmevalkit}.
We benchmark both proprietary and open-source VLMs spanning different architectural families. 
The proprietary models include \textit{Gemini~2.5~Pro}~\citep{comanici2025gemini}, \textit{GPT-5}~\citep{gpt5}, and \textit{Claude~Sonnet-4.5}~\citep{claude}. 
For open-source systems, we cover two capacity tiers. 
At the larger model size, we evaluate \textit{Qwen3-VL-32B}~\citep{qwen3technicalreport} and \textit{InternVL3.5-30B-A3B}~\citep{wang2025internvl3_5}. 
At the 7B to 8B model size, we include \textit{Qwen3-VL-8B}~\citep{qwen3technicalreport}, \textit{InternVL3.5-8B}~\citep{wang2025internvl3_5}, \textit{LLaVA-Next-Llama3}~\citep{liu2024llavanext}, \textit{LLaVA-Next-Vicuna-7B}~\citep{liu2024llavanext}, \textit{LLaVA-Next-Interleave-7B}~\citep{liu2024llavanext}, and \textit{LLaVA-OneVision-7B}~\citep{liu2024llavanext}. 
All models are evaluated using explicit, task-specific instructions. As described in \cref{sec:benchmark}, we use a multiple-choice format where each model selects from predefined options~\citep{tong2024cambrian}. Answer formats are validated using regular expressions or an auxiliary LLM~\citep{comanici2025gemini}. For template-based questions, accuracy is measured by exact match; for human-authored questions, responses are evaluated by annotators.


\subsection{Main Results}
 ~\cref{tab:main_results} reports model accuracy on each reasoning task We report accuracy separately on the \textit{simulation} and \textit{real-image} subsets to analyze cross-domain generalization, and in independent expert handcrafted questions.
 (1)  We observe that close-source models consistently outperform open-source models across almost all reasoning types, though the gap varies by task. Gemini-2.5 Pro achieves the highest overall accuracy on real images (\textbf{77.97\%}) and excels particularly in \textit{attribute matching} (\textbf{81.0\%}) and \textit{object description} (\textbf{92.2\%}), reflecting strong perceptual grounding. 
GPT-5 performs most robustly in the simulation set (\textbf{49.6\%} overall) and shows balanced accuracy across all reasoning categories, indicating stable domain transfer. 
(2) Among open-source models, Qwen3-VL-32B leads across all metrics, achieving the highest \textit{spatial reasoning} score (\textbf{67.2\%}), surpassing even close-source models while maintaining competitive performance on \textit{attribute matching} (\textbf{43.5\%}) and \textit{taxonomy reasoning} (\textbf{43.6\%}). 
(3) Taxonomy reasoning remains the most challenging category, with accuracies ranging from \textbf{28\% to 63\%}, significantly lower than spatial or attribute-based questions. On the expert-crafted questions, the best-performing model answers only achieves 24\% accuracy. 
This consistent deficit suggests that current multimodal models, despite strong visual recognition capabilities, lack explicit perceptual–taxonomy understanding and struggle to represent hierarchical object relationships beyond surface-level semantics.



\begin{figure*}[!t]
\centering
\includegraphics[width=0.8\linewidth]{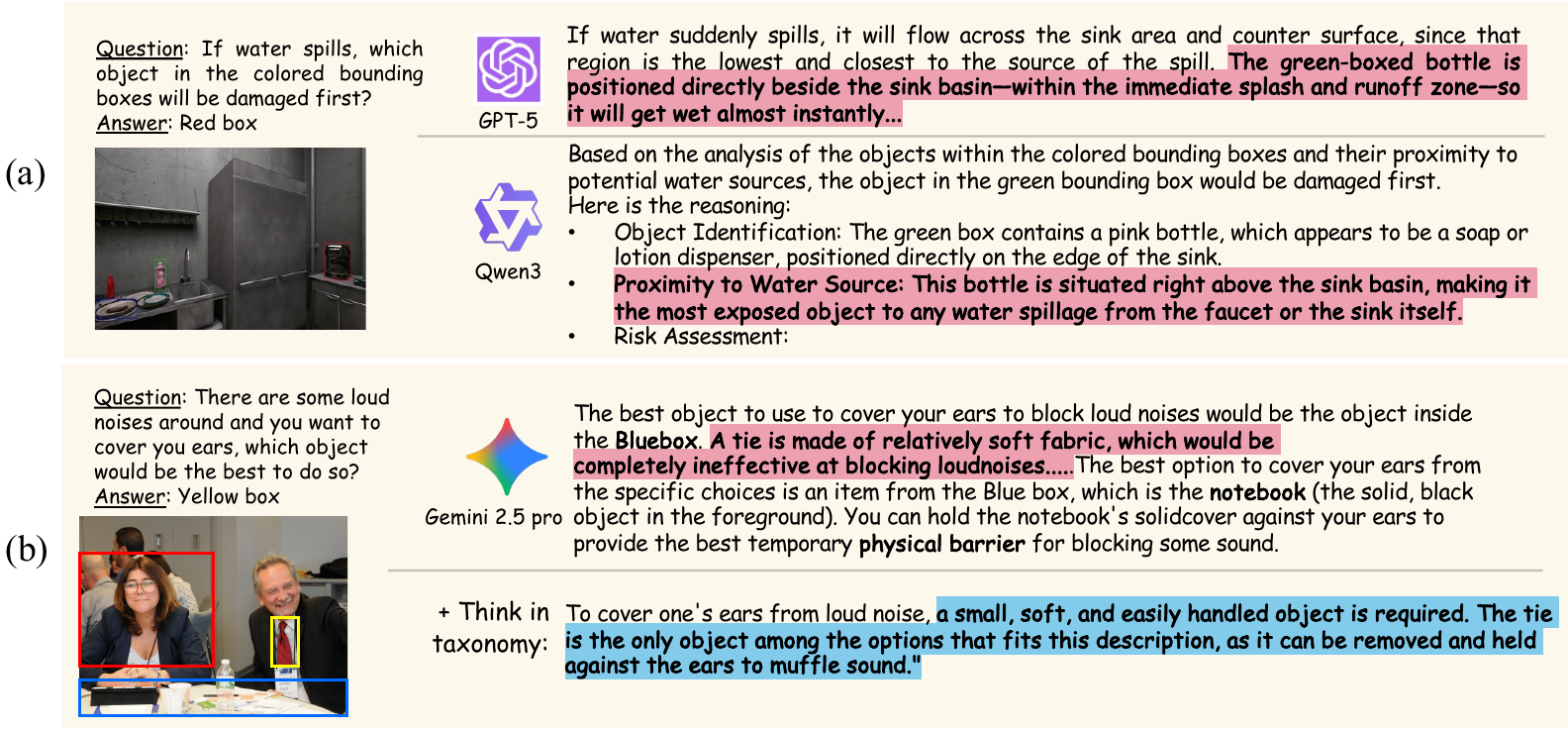}
\caption{Qualitative examples illustrating model errors in hierarchical reasoning. (a) Instances of hallucinated or inconsistent reasoning on simulated-image benchmark questions produced by Qwen3 and GPT-5. (b)  Examples of Gemini’s reasoning on real-image benchmark questions before and after in-context learning.}
\label{fig:qualitative}
\end{figure*}

\subsection{Perceptual Taxonomic Reasoning with In-Context Learning}

\begin{table}[!t]
\centering
\caption{
\small \textbf{Sim to Real improvement via taxonomic in context learning}. Providing taxonomic reasoning examples from simulation improves real image performance, especially for instruction tuned models like Gemini-2.5-Pro and GPT-5, with notable gains also seen in open source models such as InternVL3.5-8B.
}
\label{tab:icl_results}
\resizebox{\linewidth}{!}{
\begin{tabular}{l|cccc|c}
\toprule
\multirow{2}{*}{\textbf{Model}} 
& \multicolumn{4}{c|}{\textbf{Real Image Set--Taxonomy Reasoning}} 
& \multirow{2}{*}{\textbf{Expert Q.}} \\ 
\cmidrule(lr){2-5}
& \textbf{Physical} & \textbf{Material} & \textbf{Function} & \textbf{Avg.} 
&  \\ 
\midrule
\multicolumn{6}{c}{\textit{Close-source Models}} \\
\midrule
Gemini 2.5 Pro & 64.75 & 53.02 & 67.24 & 63.67 & 14.00 \\
\rowcolor{blue!10}
Gemini 2.5 Pro (w/ \dataset~ ICL) & \textbf{65.37} & \textbf{53.38} & \textbf{70.98} & \textbf{65.90} & \textbf{34.00} \\
\midrule
GPT-5 & 60.49 & 51.25 & 59.48 & 59.18 & 20.00 \\
\rowcolor{blue!10}
GPT-5 (w/ \dataset~ ICL) & \textbf{66.17} & \textbf{52.49} & \textbf{67.53} & \textbf{64.78} & \textbf{30.00}\\
\midrule
\multicolumn{6}{c}{\textit{Open-source Models}} \\
\midrule
InternVL3.5-8B & 36.91 & 37.37 & 41.95 & 37.75 & 16.00 \\
\rowcolor{blue!10}
InternVL3.5-8B (w/ \dataset~ ICL) & \textbf{48.33} & \textbf{39.15} & \textbf{43.10} & \textbf{46.38} & \textbf{25.00} \\
\midrule
Qwen3-VL-8B & 43.95 & \textbf{45.91} & \textbf{50.86} & 45.26 & 24.00 \\
\rowcolor{blue!10}
Qwen3-VL-8B (w/ \dataset~ ICL) & \textbf{46.73} & 44.13 & 47.70 & \textbf{46.55} & \textbf{31.00} \\
\bottomrule
\end{tabular}}
\end{table}

To validate the importance of perceptual taxonomy, we design a \textbf{hierarchical reasoning paradigm} that guides the model from scene layout to object-level physical properties before answering. We provide one simulation exemplar with structured object–property cues for in-context learning (ICL). This taxonomy-aware prompting yields clear gains on real-image tasks, demonstrating sim-to-real transfer of structured physical knowledge. We evaluate Gemini2.5 pro, GPT-5, Qwen3-VL-8B, and InternVL3.5-8B on taxonomy reasoning and expert crafted questions, using visually disjoint real images to isolate the effect of reasoning.

The results in \cref{tab:icl_results} report both overall and per-category accuracy on taxonomy reasoning and expert-crafted questions, showing consistent performance gains across both proprietary and open-source models when applying \dataset{} in an ICL setting. Models demonstrate improved recognition of object properties and reasoning about physical functions in real images, confirming that hierarchical reasoning learned from simulation can transfer effectively to real-world scenarios.

Notably, \textbf{Gemini~2.5~Pro} and \textbf{GPT-5} show the largest overall gains: Gemini improves from 63.67\% to 65.90\% in taxonomy reasoning and from 14 to 34 correct answers on expert questions; GPT-5 improves from 59.18\% to 64.78\% and from 20 to 30 correct. These results indicate that large, instruction-tuned VLMs can effectively follow hierarchical exemplars and apply taxonomic constraints during inference.
Still, \textbf{Qwen3-VL-8B} shows only marginal improvements (e.g. 45.26\% to 46.5\% average), likely due to weaker instruction-following capability and failure to fully execute the multi-step reasoning pattern in context.


\subsection{Qualitative Results}
As shown in Fig.~\ref{fig:qualitative}, models frequently produce fluent yet shallow explanations on taxonomy‐dependent questions. 
In (a), GPT‐5 and Qwen3 base their reasoning on spatial heuristics  (e.g. proximity to the sink) while neglecting material and structural cues that determine vulnerability to water, leading to plausible but incorrect answers. 
In (b), Gemini~2.5 Pro exhibits a similar error, prioritizing rigidity over material suitability when reasoning about sound insulation. 
These cases reveal a broader trend: models excel at perceptual grounding and spatial localization but fail to integrate hierarchical object knowledge, often conflating surface appearance with functional taxonomy. 
When guided by \dataset{} in‐context exemplars, reasoning becomes more structured, explicitly referencing material softness, size, and affordance, indicating emerging hierarchical understanding. 
Overall, the qualitative patterns mirror the quantitative results: current VLMs exhibit strong perceptual grounding but weak integration of taxonomic and material knowledge, limiting their ability to reason about object properties and usage.

%% file: sec/6_conclusion.tex

\vspace{-0.5em}
\section{Conclusion}

Our findings show that current VLMs primarily rely on associative pattern matching rather than explicit reasoning over object and property relationships. Even with strong textual priors, they struggle to maintain consistency under scene variations or counterfactual edits. These limitations highlight the need for structure aware objectives and inductive biases that reflect hierarchical scene representations.
With \dataset{}, we present the first benchmark that systematically evaluates \emph{perceptual taxonomy reasoning}, linking visual input to object level attributes such as material, affordance, function, and physical properties. Through large scale annotation and cross domain question generation, \dataset{} exposes consistent weaknesses in property level reasoning and enables few shot strategies such as \textit{PercepTax ICL}, which improve generalization from simulation to real world cases. 
We hope this work advances the development of VLMs with deeper, physically grounded understanding beyond superficial alignment.

%% file: sec/7_suppl.tex
\newpage
\appendix
\onecolumn
\section*{Appendix}

\input{sec/7_1_appendix_table_of_content}
\newpage

\setcounter{section}{0}
\renewcommand{\thesection}{\Alph{section}}

\section{Details of \dataset{} Benchmark}\label{app:details}
\subsection{Object Property Annotations}
\label{sec:object_property_annotations}
\normalsize
As detailed in Section 3.1 of the main paper, we establish a comprehensive taxonomy encompassing 16 material categories, 43 affordance types, 23 functional classes, and 17 physical properties, as depicted in Figure 2(b). Through this framework, we annotated a total of 3,173 distinct objects across both simulated and real-world scenes.

Table~\ref{tab:object_taxonomy_mapping} presents representative examples demonstrating how objects are systematically mapped to their corresponding taxonomy clusters across the four attribute domains (material, physical property, function, and affordance), accompanied by concise descriptive summaries for each object.

\setlength{\tabcolsep}{3pt}
\renewcommand{\arraystretch}{1.05}

\begin{longtable}{
  >{\raggedright\arraybackslash}p{1.8cm}
  !{\vrule width 0.6pt}
  >{\raggedright\arraybackslash}p{2.1cm}
  !{\vrule width 0.6pt}
  >{\raggedright\arraybackslash}p{2.1cm}
  !{\vrule width 0.6pt}
  >{\raggedright\arraybackslash}p{2.1cm}
  !{\vrule width 0.6pt}
  >{\raggedright\arraybackslash}p{2.1cm}
  !{\vrule width 0.6pt}
  >{\raggedright\arraybackslash}p{3.6cm}
}
\caption{Ten example objects with their taxonomy cluster mappings across material,
physical property, function, and affordance domains, along with concise descriptions.}
\label{tab:object_taxonomy_mapping}\\

\toprule
\textbf{Object} &
\textbf{Material} &
\textbf{Physical Property} &
\textbf{Function} &
\textbf{Affordance} &
\textbf{Description} \\
\midrule
\endfirsthead

\toprule
\textbf{Object} &
\textbf{Material} &
\textbf{Physical Property} &
\textbf{Function} &
\textbf{Affordance} &
\textbf{Description} \\
\midrule
\endhead

\bottomrule
\endlastfoot

\rowcolor{gray!2}
Anemone &
Biological (Plants or Flowers) &
Light, Rigid, Fragile, Movable, Stable, Solid-core, Smooth, Solid &
Botanical, Wildlife, or Environmental Support &
Grow or Plant (Vegetation) &
A sessile marine invertebrate with a cylindrical body and crown of tentacles. \\
\midrule
Ashtray &
Ceramics, Porcelain and Earthenware &
Light, Rigid, Movable, Stable, Hollow, Durable, Smooth, Solid &
Containers, Vessels, Bags, or Holding &
Contain, Carry or Package &
A receptacle designed to collect cigarette ash. \\
\midrule
\rowcolor{gray!2}
Basin &
Ceramics, Porcelain and Earthenware &
Light, Rigid, Movable, Stable, Solid-core, Durable, Smooth, Solid &
Cleaning and Sanitation &
Household or Facility Operations &
A shallow, bowl-shaped container typically used for washing. \\
\midrule
Bath Mirror &
Glass and Transparent (Silicate) &
Light, Rigid, Fragile, Movable, Stable, Solid-core, Smooth, Solid &
Viewing, Presentation, or Signage &
Display, Exhibit, Signal Value &
A reflective surface designed for personal grooming. \\
\midrule
\rowcolor{gray!2}
Beer &
Liquids and Semi-liquids &
Liquid &
Food or Drink &
Food or Prepared Dishes &
An alcoholic beverage made from fermented grains, hops, and yeast. \\
\midrule
Adjustable TV Cart &
Metals and Alloys &
Heavy, Rigid, Movable, Stable, Solid-core, Durable, Smooth, Solid &
Furniture, Storage, or Interiors &
Operate or Use Device &
A wheeled stand designed to hold and position a television. \\
\midrule
\rowcolor{gray!2}
Acorn & Organic Food and Edible Matter &
Light, Rigid, Movable, Stable, Solid-core, Durable, Smooth, Solid &
Botanical, Wildlife, or Environmental Support &
Grow or Plant (Vegetation) &
A nut from an oak tree, serving as food for wildlife. \\
\midrule


Book & Paper, Cardboard and Pulp &
Light, Rigid, Movable, Stable, Solid-core, Durable, Smooth, Solid &
Documents, Writing, Office, or Education &
Mediated Action and Meaning &
A collection of written or printed pages bound together. \\
\midrule
\rowcolor{gray!2}
Table &
Wood and Plant-based Solids &
Heavy, Rigid, Movable, Stable, Solid-core, Durable, Smooth, Solid &
Furniture, Storage, or Interiors &
Furniture, Place or Support or Work On &
A flat-topped furniture piece used for placing items or dining. \\
\midrule
T-shirt &
Textiles, Fibers and Leather &
Light, Flexible, Movable, Stable, Solid-core, Durable, Smooth, Solid &
Clothing, Soft Goods, or Human Comfort &
Wearables and Apparel &
A short-sleeved garment made of fabric, typically cotton. \\
\end{longtable}

\FloatBarrier
\clearpage

\subsection{All question templates and examples}
\label{sec:qa_templates}
Question templates grouped by category, showing diverse reasoning types across description, attribute matching (material, function, affordance, and physical property), taxonomy reasoning, and spatial relations. Placeholders  \{description\} denote object description from object annotations.

\normalsize
\setlength{\tabcolsep}{4pt}
\renewcommand{\arraystretch}{0.95}
\setlength{\LTcapwidth}{\textwidth}
\setlength{\LTpre}{0pt}
\setlength{\LTpost}{5mm}

\begin{longtable}{@{}p{0.26\textwidth} p{0.70\textwidth}@{}}
\caption{Question templates grouped by category, showing diverse reasoning types across description, attribute matching (material, function, affordance, and physical property), taxonomy reasoning, and spatial relations. Placeholders  \{description\} denote object description from object annotations.}
\label{tab:qa_templates}\\

\toprule
\textbf{Question Type} & \textbf{Example Template} \\
\midrule
\endfirsthead

\toprule
\textbf{Question Type} & \textbf{Example Template} \\
\midrule
\endhead

\bottomrule
\endlastfoot

\multicolumn{2}{l}{\textbf{Object Description}}\\
\midrule
\hspace{1em}Description match & Which object matches this description: \emph{\{description\}}? \\
\midrule
\multicolumn{2}{l}{\textbf{Attribute Matching -- Material}}\\
\midrule
\hspace{1em}Animals or Body Part & Which object is made from biological animal parts or tissues? \\
\midrule
\hspace{1em}Plants or Flowers & Which object is made from plant or floral material? \\
\midrule
\hspace{1em}\makecell[l]{Ceramics, Porcelain, \\Earthenware} & Which object is made of ceramic, porcelain, or earthenware? \\
\midrule
\hspace{1em}Gases, Vapors, Atmospheric & Which object contains or emits gases or vapors? \\
\midrule
\hspace{1em}Glass and Transparent & Which object is made of transparent or glass-like material? \\
\midrule
\hspace{1em}\makecell[l]{Glass and Transparent \\(Silicate)} & Which object is made of silicate-based transparent glass? \\
\midrule
\hspace{1em}Liquids and Semi-liquids & Which object is composed primarily of liquids or semi-liquids? \\
\midrule
\hspace{1em}Metals and Alloys & Which object is made of metallic materials or alloys? \\
\midrule
\hspace{1em}\makecell[l]{Organic Food and \\Edible Matter} & Which object is made of organic edible material or food matter? \\
\midrule
\hspace{1em}Paper, Cardboard and Pulp & Which object is made of paper, cardboard, or pulp? \\
\midrule
\hspace{1em}\makecell[l]{Plastics, Rubber and \\Polymers} & Which object is made of plastic, rubber, or polymer material? \\
\midrule
\hspace{1em}Stone, Concrete and Mineral & Which object is made of stone, concrete, or mineral material? \\
\midrule
\hspace{1em}Textiles, Fibers and Leather & Which object is made of textile, fiber, or leather materials? \\
\midrule
\hspace{1em}Wood and Plant-Based Solids & Which object is made of wood or other plant-based solid materials? \\
\midrule

\multicolumn{2}{l}{\textbf{Attribute Matching -- Function}}\\
\midrule
\hspace{1em}Functional knowledge & Which object is used as ``a container''? \\
\midrule
\hspace{1em}Tableware or serveware & Which object has the affordance of tableware or serveware? \\
\midrule
\hspace{1em}Food or produce & Which object is food or produce? \\
\midrule
\hspace{1em}Prepared food & Which object is prepared food? \\
\midrule
\hspace{1em}Furniture & Which of these objects is likely to be considered furniture? \\
\midrule
\hspace{1em}Architectural fixture & Which object functions as an architectural component or fixture? \\
\midrule
\hspace{1em}Structural element & Which object functions as a structural element that spans or occupies space? \\
\midrule
\hspace{1em}Plant support & Which object functions as a holder or support for growing plants? \\
\midrule
\hspace{1em}Household or facility item & Which object supports household or facility operations? \\
\midrule
\hspace{1em}Machine or appliance & Which object functions as a powered machine or appliance? \\
\midrule
\hspace{1em}\makecell[l]{Media or communication \\device} & Which object functions as a device for reading, writing, or communication? \\
\midrule
\hspace{1em}Seating or riding object & Which object functions as a seat or item for riding? \\
\midrule
\hspace{1em}Interaction with organisms & Which object is used to interact with living or moving things? \\
\midrule
\multicolumn{2}{l}{\textbf{Attribute Matching -- Affordance}}\\
\midrule
\hspace{1em}Container or carrier & Which object looks like it could hold or carry other items inside it? \\
\midrule
\hspace{1em}Cleaning or sanitation & Which object seems intended for cleaning or wiping surfaces? \\
\midrule
\hspace{1em}Light source or control & Which object could produce or control light in this scene? \\
\midrule
\hspace{1em}Display or signal device & Which object could display or signal information or value? \\
\midrule
\hspace{1em}Art display & Which object has the affordance of art display (view or appraise)? \\
\midrule
\hspace{1em}Handheld or portable object & Which object seems small or shaped for someone to grip and carry easily? \\
\midrule
\hspace{1em}Manual tool or device & Which object could be operated by hand without needing electricity or power? \\
\midrule
\hspace{1em}Work surface & Which object seems flat and sturdy enough to place items on? \\
\midrule
\hspace{1em}Wearable item & Which object has the affordance of wearables or apparel? \\
\midrule
\hspace{1em}Enclosure or shelter & Which object appears to be an enclosed space or shelter someone could enter? \\
\midrule

\multicolumn{2}{l}{\textbf{Attribute Matching -- Physical Property}}\\
\midrule
\hspace{1em}Durable & Which object is durable and resistant to wear or damage? \\
\midrule
\hspace{1em}Fixed & Which object appears fixed in place and not easily movable? \\
\midrule
\hspace{1em}Flexible & Which object is flexible and can bend without breaking? \\
\midrule
\hspace{1em}Fragile & Which object is fragile and could break easily? \\
\midrule
\hspace{1em}Gas & Which object exists in gaseous form or emits gas? \\
\midrule
\hspace{1em}Heavy & Which object appears heavy or dense? \\
\midrule
\hspace{1em}Hollow & Which object appears hollow inside? \\
\midrule
\hspace{1em}Light & Which object is lightweight and easy to move? \\
\midrule
\hspace{1em}Liquid & Which object is in a liquid state? \\
\midrule
\hspace{1em}Movable & Which object can be moved easily from one place to another? \\
\midrule
\hspace{1em}Rigid & Which object is rigid and maintains a fixed shape? \\
\midrule
\hspace{1em}Rough & Which object has a rough or uneven surface texture? \\
\midrule
\hspace{1em}Smooth & Which object has a smooth surface texture? \\
\midrule
\hspace{1em}Solid & Which object is solid and not hollow? \\
\midrule
\hspace{1em}Solid-core & Which object has a solid internal core? \\
\midrule
\hspace{1em}Stable & Which object is stable and unlikely to tip or fall? \\
\midrule
\hspace{1em}Unstable & Which object appears unstable or easily tipped over? \\
\midrule

\multicolumn{2}{l}{\textbf{Taxonomy Reasoning -- Physical}}\\
\midrule
\hspace{1em}Sturdy non-container & Which object is rigid, movable, but NOT designed as a container? (Exclude living organisms.) \\
\midrule
\hspace{1em}Hidden storage & Which object can hide small items while keeping the area tidy? \\
\midrule
\hspace{1em}Compressible object & Which object can be compressed to fit in tight spaces without damaging its structure? \\
\midrule

\multicolumn{2}{l}{\textbf{Taxonomy Reasoning -- Material}}\\
\midrule
\hspace{1em}Heat-sensitive object & Which object would be most affected by high heat? \\
\midrule
\hspace{1em}Water-sensitive object & If water spills, which object gets damaged first? \\
\midrule
\hspace{1em}Scratch-resistant surface & Which surface is least likely to show scratch marks if you scratch it with a fingernail? \\
\midrule
\hspace{1em}Sound-absorbing object & There are loud noises around-- which object would best cover your ears? \\
\midrule
\hspace{1em}Cold-to-touch material & Which object would feel coldest to touch in a cold room? \\
\midrule

\multicolumn{2}{l}{\textbf{Taxonomy Reasoning -- Function}}\\
\midrule
\hspace{1em}Foldable object & Which object can be folded or collapsed to save space? \\
\midrule
\hspace{1em}Reuse as shield & What object in the scene could be repurposed as a shield? \\
\midrule
\hspace{1em}Reuse as container & If no bag or box is available, which object could temporarily hold or transport items? \\
\midrule
\hspace{1em}Reuse as reflector & Imagine sunlight or a flashlight shining--what item in the scene would best reflect that light? \\
\midrule
\hspace{1em}Reuse as cushion & If someone wants a temporary pillow, which object could be folded or rolled up for cushioning? \\
\midrule
\hspace{1em}Reuse as stepstool & If you had to reach a higher shelf, which object could double as a stepstool? \\
\midrule
\hspace{1em}Reuse as bookend & Imagine organizing a shelf without real bookends--what nearby item could hold books in place? \\
\midrule

\multicolumn{2}{l}{\textbf{2D/3D Spatial Relations}}\\
\midrule
\hspace{1em}Spatial: above or below & Is the object marked by the Red box above or below the object marked by the Green box? \\
\midrule
\hspace{1em}Spatial: closer or farther & Is the object marked by the Red box or the Green box closer to the camera? \\
\midrule
\hspace{1em}Spatial: front or behind & Is the object marked by the Red box in front of or behind the object marked by the Green box? \\
\midrule
\hspace{1em}Spatial: left or right & Is the object marked by the Green box to the left or right of the object marked by the Blue box? \\
\end{longtable}

\newpage
\subsection{Question Type Statistics}
\label{sec:qa_category_stats}
\FloatBarrier

\begin{table}[!htbp]
\caption{Aggregated question statistics across benchmark tasks and question categories for simulated (Sim) and real-image (Real) benchmarks. }
\centering
\small
\setlength{\tabcolsep}{5pt}
\renewcommand{\arraystretch}{0.95}

\begin{tabular}{lrrr}
\toprule
\textbf{Category} & \textbf{Sim (4{,}544 scenes)} & \textbf{Real (1{,}258 photos)} & \textbf{Total} \\
\midrule
\multicolumn{4}{l}{\textbf{Object Description}}\\
\hspace{1em}Object Description & 236 & 255 & 491 \\
\midrule
\textbf{Subtotal -- Object Description} & \textbf{236} & \textbf{255} & \textbf{491} \\[3pt]
\midrule
\multicolumn{4}{l}{\textbf{Attribute Matching}}\\
\hspace{1em}Material & 108 & 232 & 340 \\
\hspace{1em}Function & 1{,}293 & 869 & 2{,}162 \\
\hspace{1em}Physical & 120 & 19 & 139 \\
\hspace{1em}Affordance & 2{,}338 & 1{,}213 & 3{,}551 \\
\midrule
\textbf{Subtotal -- Attribute Matching} & \textbf{3{,}860} & \textbf{2{,}333} & \textbf{6{,}193} \\[3pt]
\midrule
\multicolumn{4}{l}{\textbf{Taxonomy Reasoning}}\\
\hspace{1em}Physical & 2{,}189 & 731 & 2{,}920 \\
\hspace{1em}Material & 4{,}130 & 585 & 4{,}715 \\
\hspace{1em}Function & 7{,}252 & 1{,}018 & 8{,}270 \\
\midrule
\textbf{Subtotal -- Taxonomy Reasoning} & \textbf{13{,}571} & \textbf{2{,}334} & \textbf{15{,}905} \\[3pt]
\midrule
\multicolumn{4}{l}{\textbf{2D/3D Spatial Relations}}\\
\hspace{1em}Spatial: above or below & 1{,}231 & 226 & 1{,}457 \\
\hspace{1em}Spatial: closer or farther & 1{,}263 & 0 & 1{,}263 \\
\hspace{1em}Spatial: front or behind & 1{,}263 & 145 & 1{,}408 \\
\hspace{1em}Spatial: left or right & 1{,}221 & 96 & 1{,}317 \\
\midrule
\textbf{Subtotal -- 2D/3D Spatial Relations} & \textbf{4{,}978} & \textbf{467} & \textbf{5{,}445} \\[3pt]
\midrule
\multicolumn{4}{l}{\textbf{Human Expert}}\\
\hspace{1em} Human-level Reasoning  & 0 & 50 & 50 \\
\midrule
\textbf{Total (All Categories)} & \textbf{22{,}644} & \textbf{5{,}439} & \textbf{28{,}083} \\
\bottomrule
\end{tabular}
\label{tab:qa_category_stats}
\end{table}

\FloatBarrier
\newpage
\section{Operational Details of Template-Based QA Generation}
\normalsize
\label{sec:qa_details_appendix}

Our QA generation pipeline expands the 74 parameterized templates described in Section~\ref{sec:task_types} into large-scale, semantically grounded question--answer pairs for both simulation and real-image scenes. The pipeline operates on the structured per-object annotations introduced in Section~\ref{sec:scene_structure}, including material, function, affordances, physical properties, and 2D/3D spatial metadata. 

\subsection{Scene Curation}
\label{sec:scene_curation}
\noindent\textbf{Simulation Images.} We construct the synthetic image set using $33$ high-quality Unreal Engine 5 \cite{unrealengine5} assets, which offer precise geometry, physically based materials, and photorealistic illumination. To preserve meaningful visual cues such as shadows, reflections, and occlusions, we manually configure lighting and exposure settings for each scene. We then employ a two-stage automatic pipeline to sample diverse and semantically rich camera viewpoints. In the first stage, candidate camera locations are generated via Poisson-disk sampling at pre-specified heights to cover the entire scene content. The sampling bounds are calculated by the 3D bounding boxes of all objects in the scene, after discarding extreme outliers. We then apply a view-quality filter based on the segmentation mask and depth map. This filter assigns each candidate view a weighted score that combines depth and segmentation quality: the depth term penalizes overly uniform maps (e.g., flat walls) and views with large depth outliers, while the segmentation term penalizes views with too few objects or a single dominant object and requires a sufficient number of small objects. Together, these scores favor viewpoints that capture diverse object scales and semantically rich layouts. In the second stage, we further filter images to retain only those that contain foreground instances of the objects of interest from our curated object list, which is constructed by extracting all object labels from the scenes and classifying them with LLMs. Each rendered image is paired with explicit spatial annotations, including 2D instance segmentations, 2D and 3D bounding boxes, and object orientations.

\begin{figure*}[t]
\centering
\includegraphics[width=0.95\linewidth]{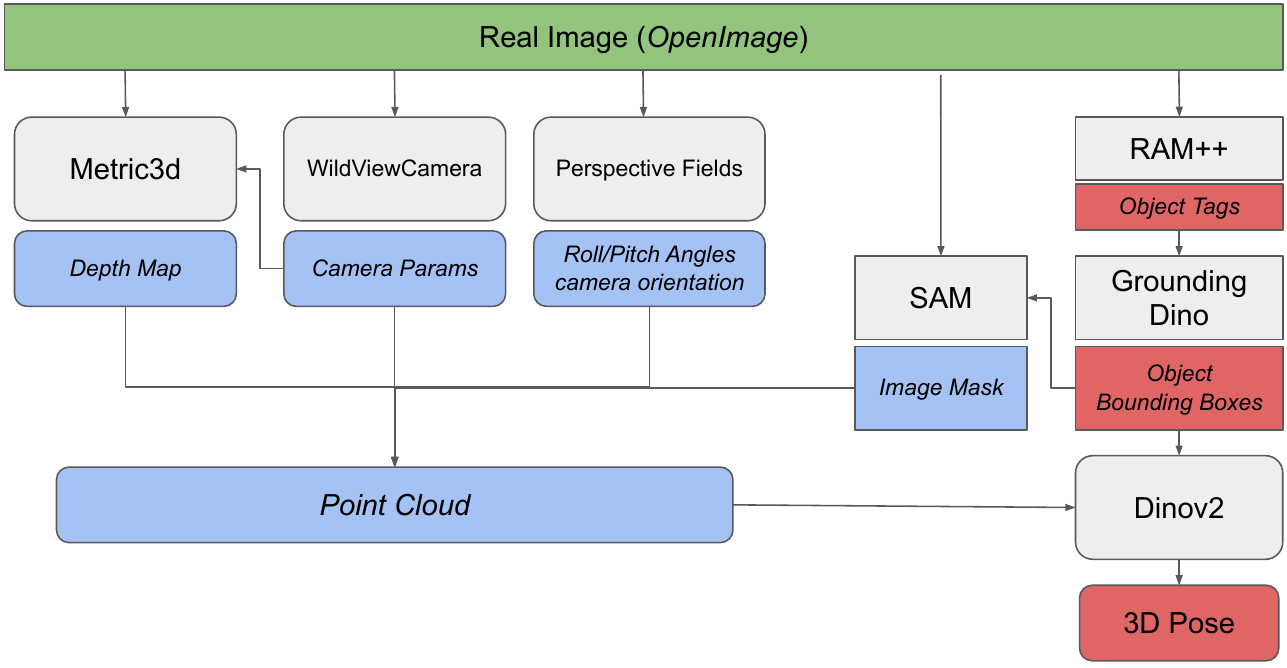}
\caption{
Overview of the Real Image object annotation pipeline the blue blocks are intermediate outputs and the red blocks (object tags, Bounding Boxes, 3D poses are the pipeline outputs we use to further create the object attribute clusters and Benchmarks Questions).
\vspace{-1em}}
\label{fig:annotation_pipeline}

\end{figure*}

\noindent\textbf{Real Image scene curation.} After curating the simulation scenes, we collected images from the OpenImages 10k training subset \cite{kuznetsova2020openimages}. We used several foundation models (e.g. RAM++ \cite{yang2024ramplusplus}, Grounding DINO \cite{liu2023groundingdino}, and DINOv2 \cite{oquab2023dinov2}) to detect objects and capture their spatial annotation data, including 2D and 3D bounding boxes and pose vectors. Pipeline overview is presented in Figure \ref{fig:annotation_pipeline}

\noindent\textbf{Object Annotation.} For each object we curated in the simulation scene and real image, we employed Gemini to generate detailed descriptive sentences encompassing visual characteristics and object attributes relevant to our taxonomy reasoning framework (e.g. material, function). These generated descriptions were then clustered using unsupervised algorithms such as HDBSCAN \cite{mcinnes2017hdbscan} and \textit{k}-means \cite{macqueen1967kmeans} to group objects with similar physical properties, materials, functions, and affordances based on the attribute descriptions produced by Gemini. Following automated clustering, human annotators manually reassigned and verified cluster memberships to ensure semantic coherence and accuracy, thereby refining the taxonomy structure. The verified clusters were then used in our QA generation pipeline, producing question–answer pairs that form the foundation of both the simulation-based and real-image benchmarks.

\subsection{Template Instantiation from Structured Annotations}
\label{sec:template_instantiation}
For each scene, we instantiate templates by binding specific cluster attribute. For instance, attribute matching questions use one specific cluster information (e.g. Attribute matching- Affordance questions ask specifically about one cluster), where as taxonomy reasoning uses multiple cluster information for each template questions) to the corresponding object attributes. The templates span four major perceptual taxonomy tasks:

\begin{itemize}
    \item \textbf{Object Description:} match a natural-language description derived from an object's attributes to the correct color-coded box.
    \item \textbf{Spatial Reasoning:} determine left/right, above/below, front/behind, and closer/farther relations based on ground-truth image coordinates and 3D camera-frame positions.
    \item \textbf{Attribute Matching:} associate specific materials, physical properties, or affordances with the correct object (1 cluster attribute).
    \item \textbf{Taxonomy Reasoning:} perform multi-step inference across multiple structured attributes (e.g., rigidity + stability + functional role).
\end{itemize}
Spatial relations are computed using 2D centers in the image plane and yaw-corrected 3D offsets in the camera coordinate frame. The resulting quantitative relations are converted into natural-language prompts.

\subsection{Attribute-Cluster Filtering}
\label{sec:attribute_filtering}
To ensure semantic validity, we filter out candidate objects whose attribute clusters are unsuitable for reasoning. Specifically,
abstract or non-physical clusters (e.g., ``depicted scene,'' ``occupations,'' or other unclassified types) are excluded from material questions. Similarly, physical-property questions omit objects whose clusters do not represent interpretable physical traits. This filtering removes cases where attribute grounding is ambiguous or not visually meaningful.

\subsection{Uniqueness and Ambiguity Suppression}
\label{sec:uniqueness_ambiguity}
We enforce a strict \textbf{single-answer guarantee}. A generated question is discarded if:
\begin{itemize}
    \item multiple objects satisfy all required attributes (e.g., several wooden objects),
    \item multiple objects share identical spatial relations under a template (e.g., equal horizontal offsets),
    \item taxonomy constraints allow more than one plausible candidate.
\end{itemize}

Only questions with exactly one valid answer are retained, ensuring clarity and preventing unintended ambiguity.

\subsection{LLM-Assisted Linguistic Diversification}
\label{sec:llm_diversification}
To enhance linguistic variety without affecting determinism, we apply lightweight LLM-based rephrasing (Gemini~\cite{comanici2025gemini}) to each template instance. Paraphrases are kept only when they preserve:
\begin{itemize}
    \item attribute grounding,
    \item bounding-box references,
    \item the original semantic intent,
    \item the uniqueness of the correct answer.
\end{itemize}

\subsection{Automatic Consistency Checks}
\label{sec:consistency_checks}
Each instantiated question undergoes a series of automated validation steps:
\begin{itemize}
    \item \textbf{Object presence check:} referenced bounding boxes must appear in the scene.
    \item \textbf{Attribute validity check:} all template attributes must match verified cluster annotations.
    \item \textbf{Parameter integrity:} template variables must be bound to valid attributes.
    \item \textbf{Answer correctness:} the computed solution must uniquely correspond to exactly one object.
\end{itemize}

Instances failing any validation are removed.

\subsection{Human Verification}
\label{sec:human_verification}
Human annotators perform targeted verification at three levels:
\begin{enumerate}
    \item \textbf{Cluster verification} after HDBSCAN/\(k\)-means grouping,
    \item \textbf{Template validation} to confirm semantic correctness,
    \item \textbf{Final QA inspection} for grammar, clarity, and visual answerability.
\end{enumerate}
All template definitions used in this pipeline are provided in Table~\ref{tab:qa_templates}.

\subsection{Final Object Annotation Counts}
\label{sec:final_counts}
Out of the \text{3{,}173} annotated objects, \text{1{,}563} come from simulation and \text{1{,}858} from real images (with overlapping objects removed from the unique count). After consistency checks, image-quality filtering, and uniqueness and ambiguity suppression, a total of \text{892} unique objects are retained for the Real-Image benchmark and \text{204} for the Simulation benchmark.
\newpage
\section{Human Level Reasoning Questions Example}
\label{sec:human_reasoning}
\FloatBarrier

\setlength{\intextsep}{0pt}      
\setlength{\textfloatsep}{0pt}   
\setlength{\floatsep}{0pt}       
\begin{figure}[H]
\caption{Here are examples of GPT-5.1 answering the human-expert questions. The blocks marked with red Xs show the model’s answers and reasoning before in-context learning using our proposed reasoning scheme, while the blocks marked with green check marks show the improved answers after applying our ICL prompting strategy.}
\centering
\includegraphics[width=0.8\linewidth]{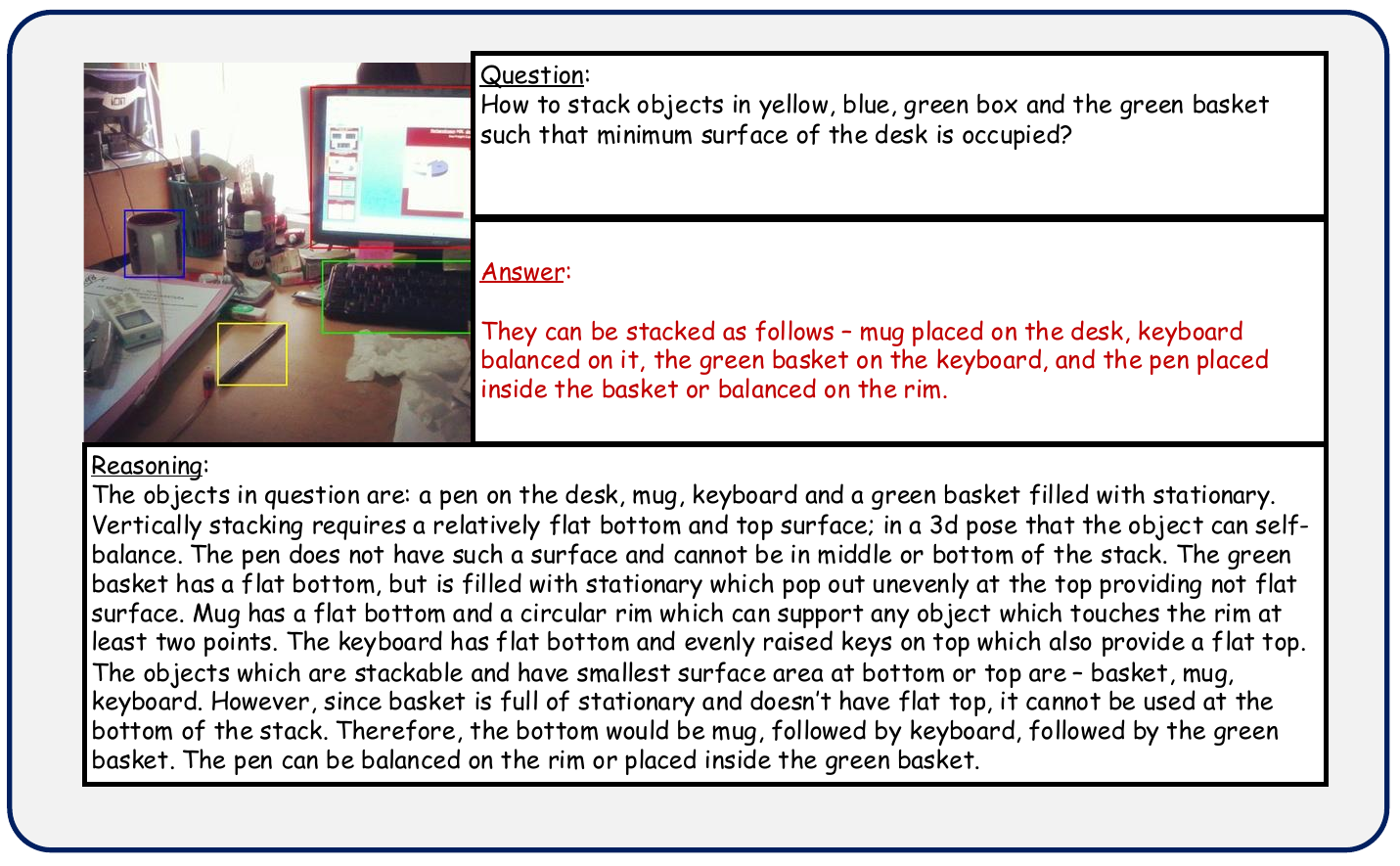}
\vspace{4pt}
\includegraphics[width=0.8\linewidth]{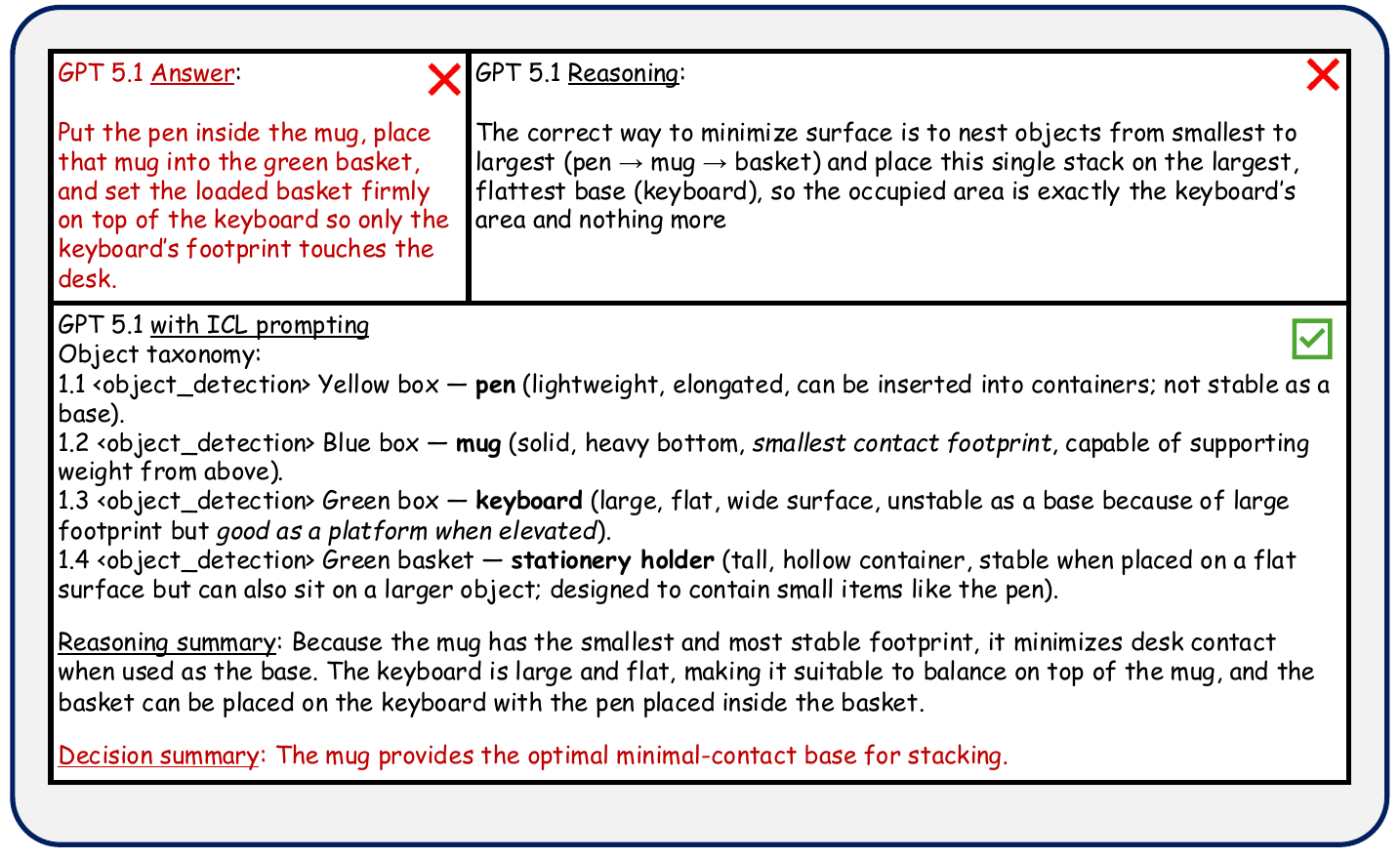}
\label{fig:qa_example}
\end{figure}

\section{Human Level Reasoning Questions Example Cont.}
\label{sec:human_reasoning_cont}
\begin{figure}[H]
\caption{Additional comparison between the original GPT-5.1 outputs and the ICL-improved answers.}
\centering
\includegraphics[width=0.8\linewidth]{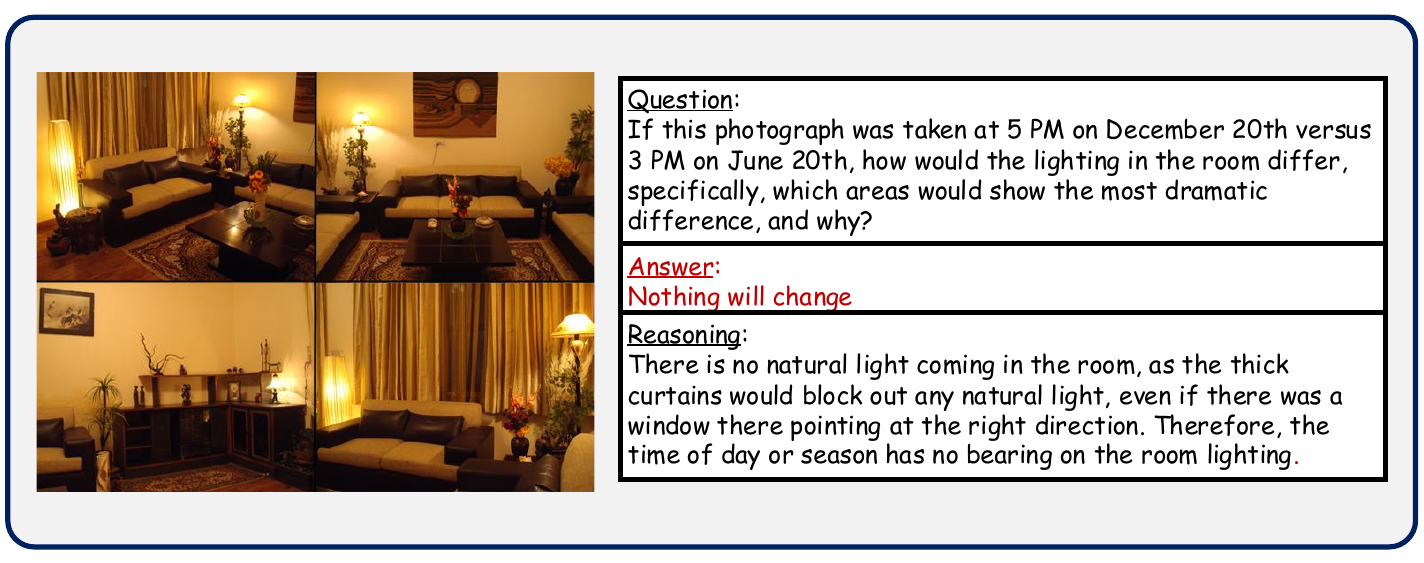}
\vspace{4pt}
\includegraphics[width=0.8\linewidth]{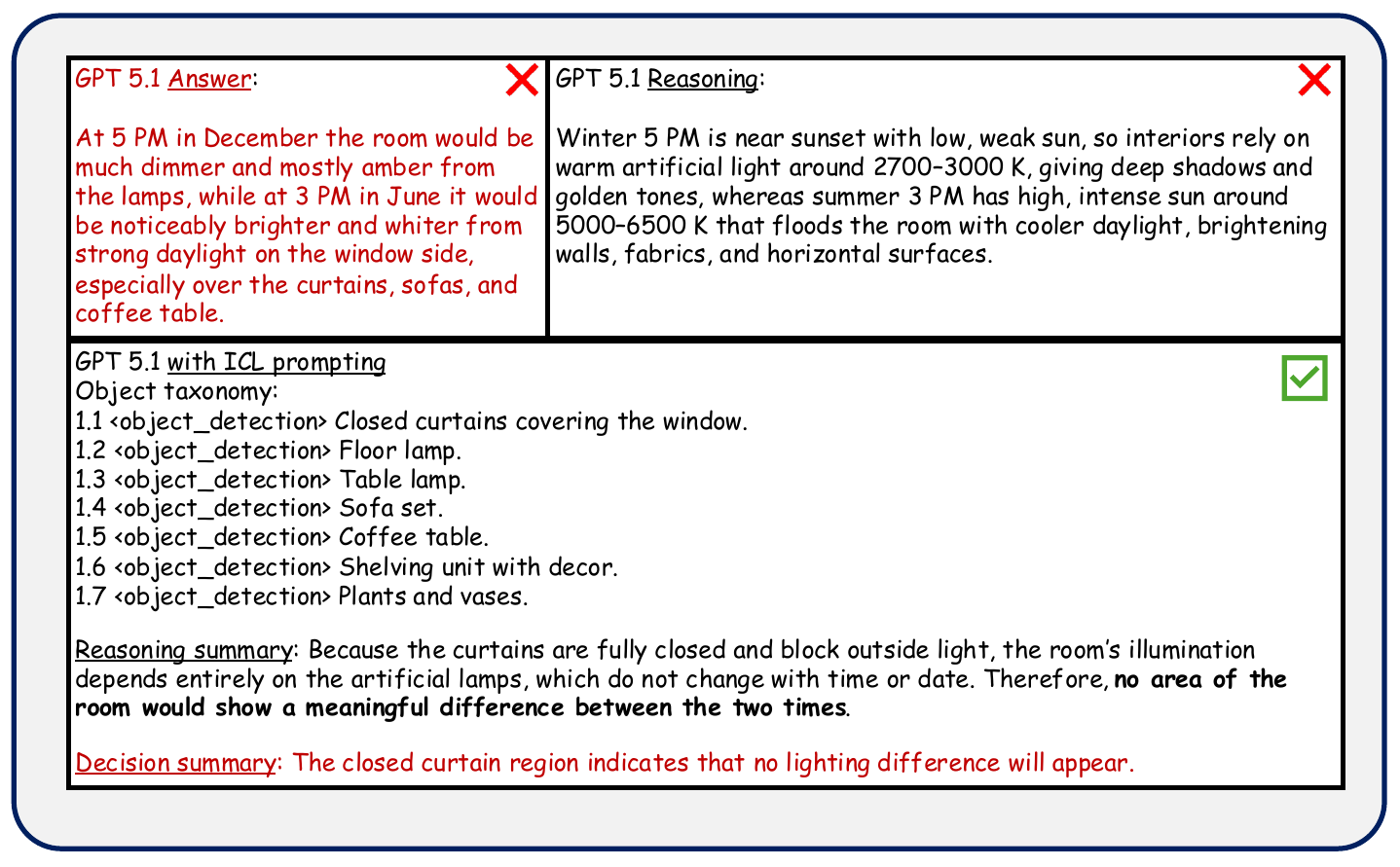}
\label{fig:qa_example}
\end{figure}

\newpage
\section{More QA Examples from the Real image Benchmark}
\label{sec:real_examples}
\FloatBarrier

\setlength{\intextsep}{0pt}      
\setlength{\textfloatsep}{0pt}   
\setlength{\floatsep}{0pt}       
\begin{figure}[H]
\centering
\includegraphics[width=0.8\linewidth]{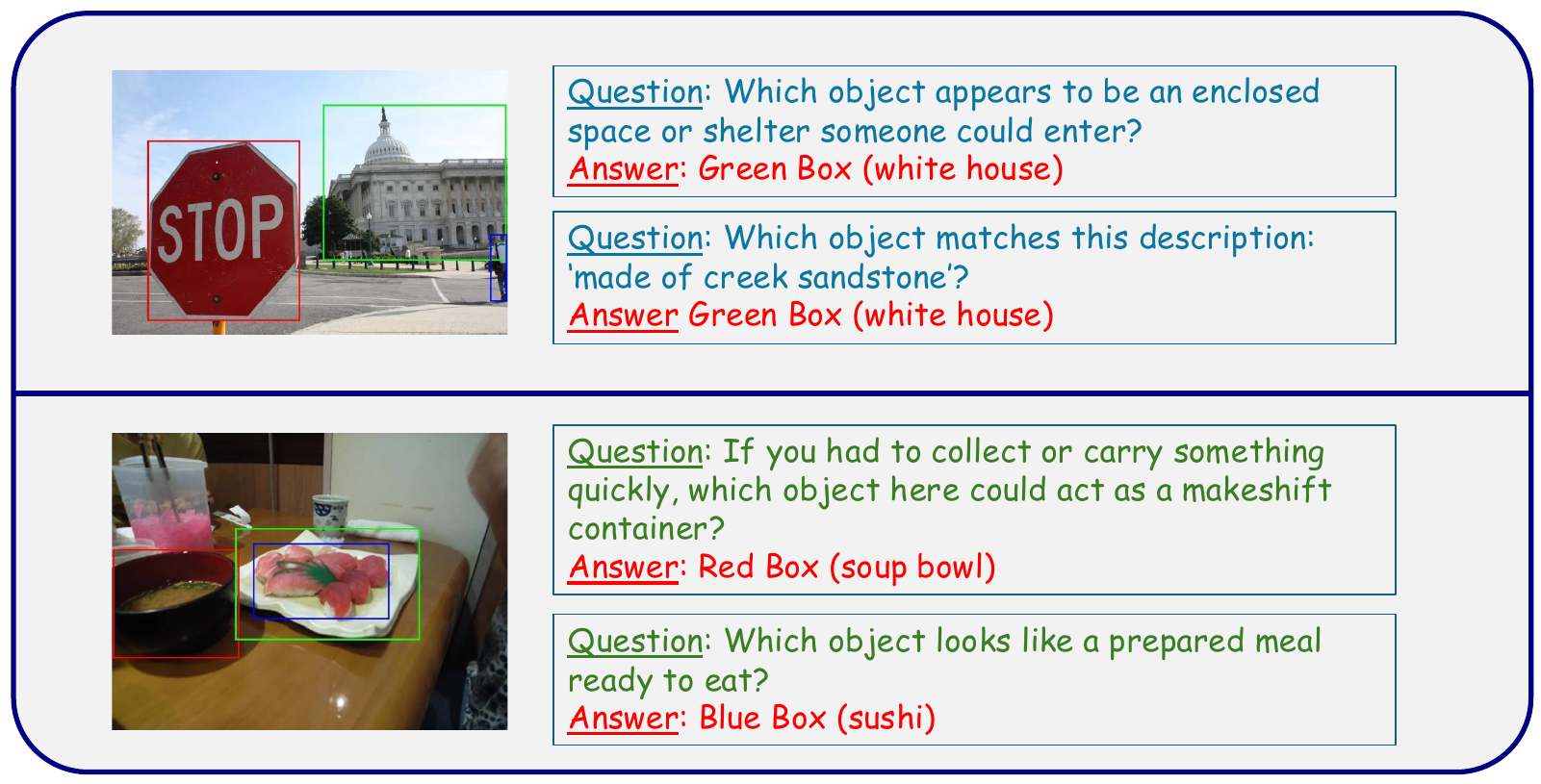}
\vspace{4pt}
\includegraphics[width=0.8\linewidth]{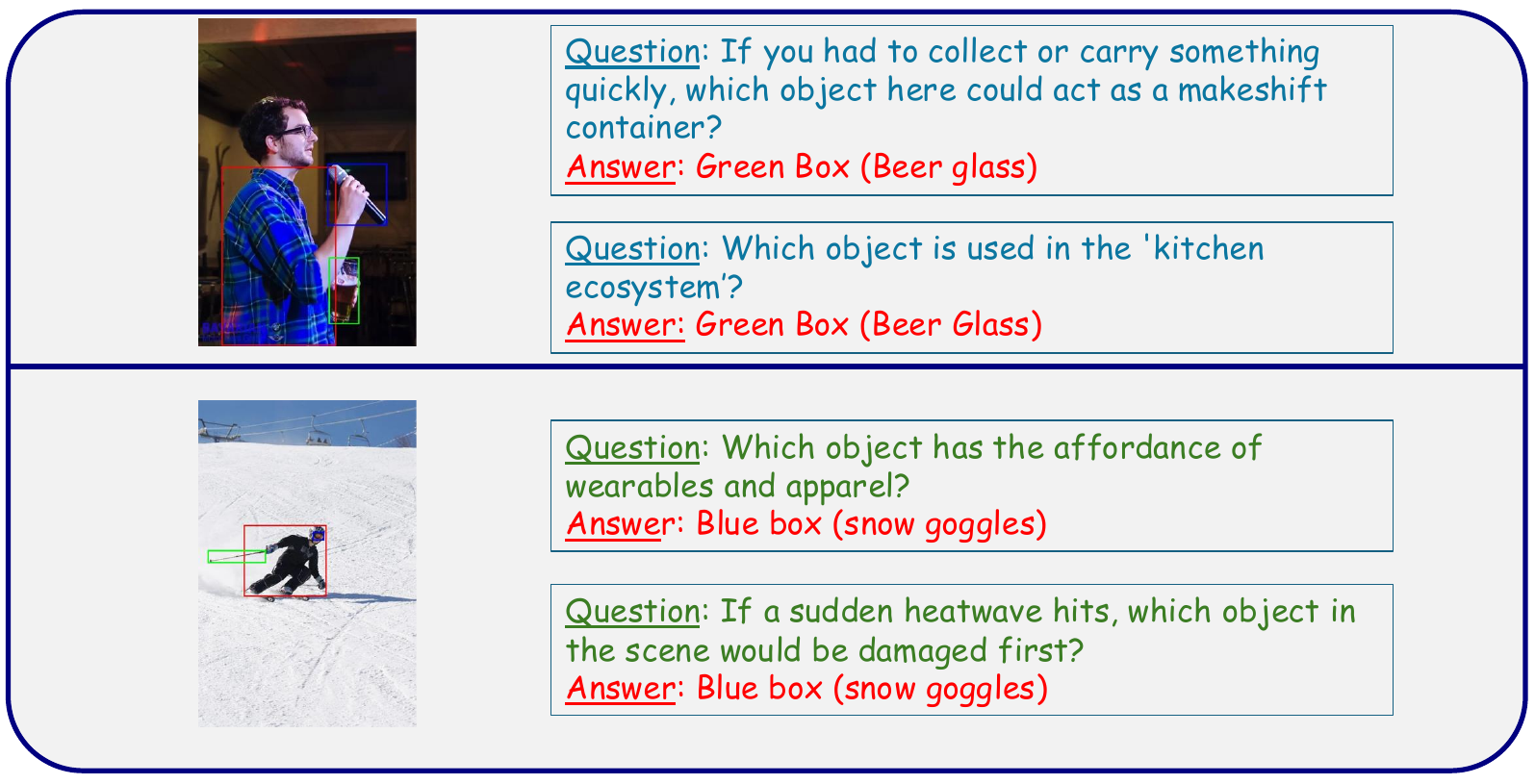}
\label{fig:qa_example}
\end{figure}

\section{More QA Examples from the Sim image Benchmark}
\label{sec:sim_examples}
\begin{figure}[H]
\centering
\includegraphics[width=0.8\linewidth]{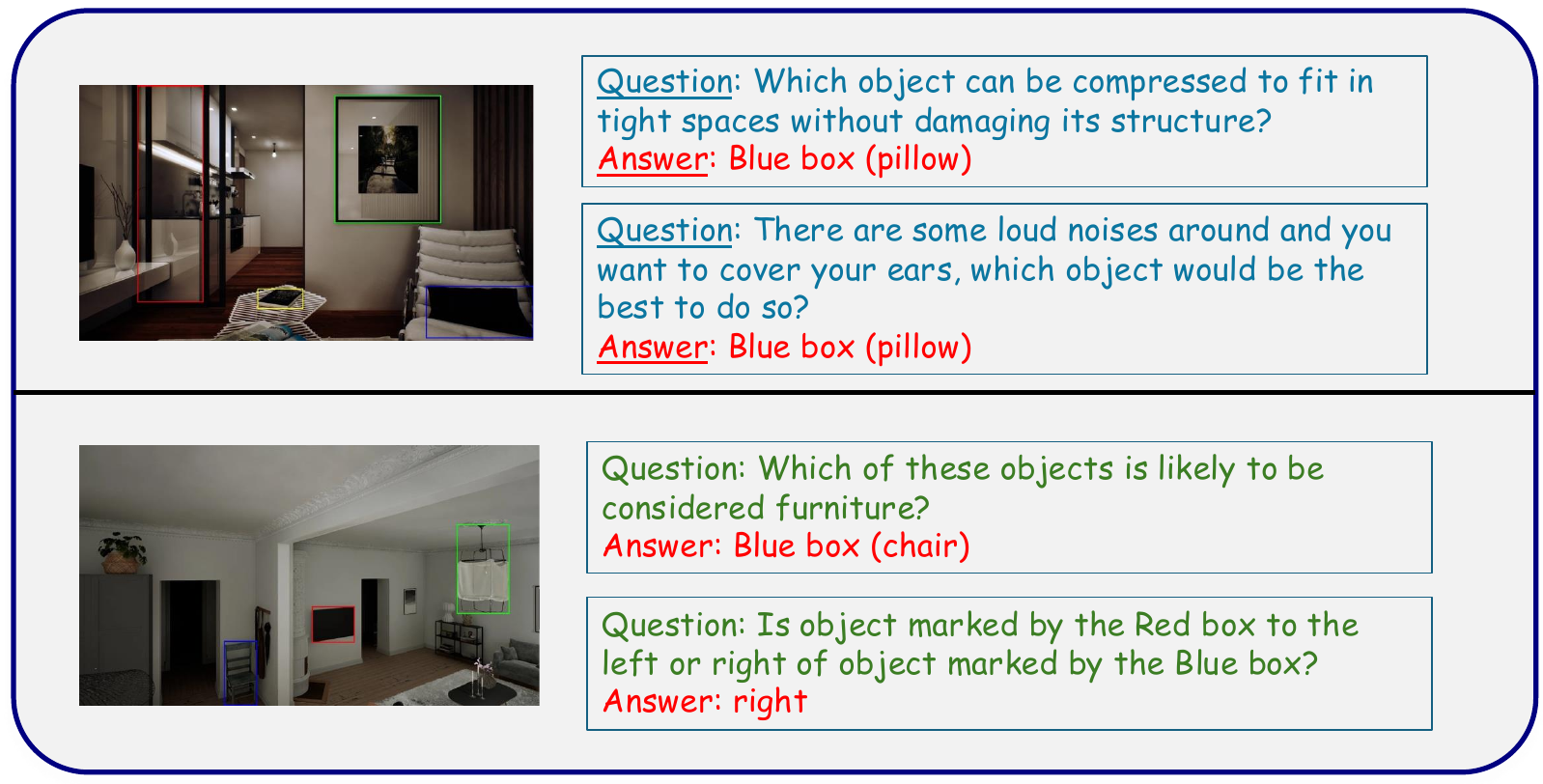}
\vspace{4pt}
\includegraphics[width=0.8\linewidth]{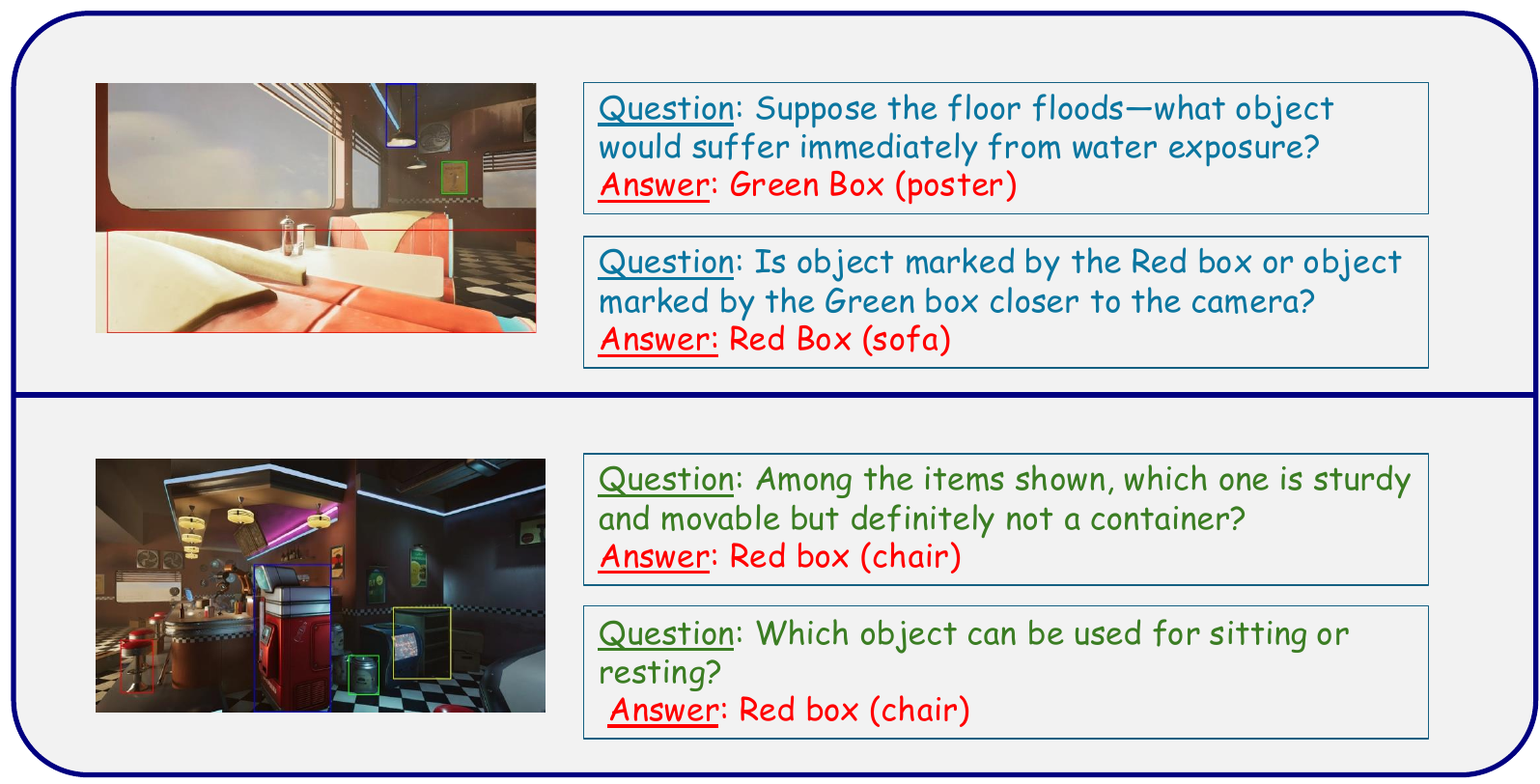}
\label{fig:qa_example}
\end{figure}

%% file: sec/7_1_appendix_table_of_content.tex
\section*{Table of Contents - Appendix}

\noindent\textbf{Appendix A. Additional Tables} \dotfill \pageref{app:details}
\begin{itemize}[leftmargin=2em, topsep=0pt, itemsep=2pt]
    \item \ref{sec:object_property_annotations}. Object Property Annotations \dotfill \pageref{sec:object_property_annotations}
    \item \ref{sec:qa_templates}. Question Templates by Category \dotfill \pageref{sec:qa_templates}
    \item \ref{sec:qa_category_stats}. Aggregated Question Statistics \dotfill \pageref{sec:qa_category_stats}
\end{itemize}

\vspace{0.5em}
\noindent\textbf{Appendix \ref{sec:qa_details_appendix}. Operational Details of Template-Based QA Generation} \dotfill \pageref{sec:qa_details_appendix}
\begin{itemize}[leftmargin=2em, topsep=0pt, itemsep=2pt]
    \item \ref{sec:scene_curation}. Scene Curation \dotfill \pageref{sec:scene_curation}
    \item \ref{sec:template_instantiation}. Template Instantiation from Structured Annotations \dotfill \pageref{sec:template_instantiation}
    \item \ref{sec:attribute_filtering}. Attribute-Cluster Filtering \dotfill \pageref{sec:attribute_filtering}
    \item \ref{sec:uniqueness_ambiguity}. Uniqueness and Ambiguity Suppression \dotfill \pageref{sec:uniqueness_ambiguity}
    \item \ref{sec:llm_diversification}. LLM-Assisted Linguistic Diversification \dotfill \pageref{sec:llm_diversification}
    \item \ref{sec:consistency_checks}. Automatic Consistency Checks \dotfill \pageref{sec:consistency_checks}
    \item \ref{sec:human_verification}. Human Verification \dotfill \pageref{sec:human_verification}
    \item \ref{sec:final_counts}. Final Object Annotation Counts \dotfill \pageref{sec:final_counts}
\end{itemize}

\vspace{0.5em}
\noindent\textbf{Appendix \ref{sec:human_reasoning}. Human Level Reasoning Questions Example} \dotfill \pageref{sec:human_reasoning}

\vspace{0.5em}
\noindent\textbf{Appendix \ref{sec:real_examples}. More QA Examples from the Real Image Benchmark} \dotfill \pageref{sec:real_examples}

\vspace{0.5em}
\noindent\textbf{Appendix \ref{sec:sim_examples}. More QA Examples from the Sim Image Benchmark} \dotfill \pageref{sec:sim_examples}